\documentclass{article} 
\PassOptionsToPackage{dvipsnames}{xcolor}
\usepackage{iclr2026_conference,times}

\usepackage{amsmath,amsfonts,bm}

\def\eqref#1{equation~\ref{#1}}

\def\1{\bm{1}}

\DeclareMathAlphabet{\mathsfit}{\encodingdefault}{\sfdefault}{m}{sl}
\SetMathAlphabet{\mathsfit}{bold}{\encodingdefault}{\sfdefault}{bx}{n}

\usepackage{hyperref}
\usepackage{url}
\usepackage{cleveref}
\usepackage{graphics} 
\usepackage{graphicx} 
\usepackage{amsmath} 
\usepackage{amssymb}  
\usepackage{natbib}
\usepackage{wrapfig}
\usepackage{booktabs}
\usepackage{multirow}

\usepackage{wrapfig}
\usepackage{tabularx}
\usepackage{caption}
\usepackage{array}
\usepackage{subcaption}

\usepackage{minitoc}
\usepackage{tocloft}
\usepackage[toc,page,header]{appendix}

\usepackage[dvipsnames]{xcolor}

\definecolor{yellow}{rgb}{1,1, 0.6}
\definecolor{lightyellow}{rgb}{1,1, 0.8}
\definecolor{orange}{rgb}{1, 0.8, 0.6}
\definecolor{red}{rgb}{1, 0.6, 0.6}
\definecolor{wincolor}{rgb}{0.85, 0.0, 0.0}
\definecolor{darkyellow}{rgb}{0.8, 0.8, 0.5}
\definecolor{darkred}{rgb}{0.7, 0.3, 0.3}
\definecolor{darkgreen}{rgb}{0.3, 0.7, 0.3}
\definecolor{blue}{rgb}{0, 0, 1.0}
\definecolor{green}{rgb}{0, 1.0, 0}
\definecolor{pink}{rgb}{1, 0.4, 0.7}

\definecolor{citecolor}{HTML}{0071bc}

\definecolor{PurpleColor}{HTML}{8B008B}
\definecolor{GreenColor}{rgb}{0.137,0.573,0.565}

\title{ReCal3R: Reliability-Calibrated Learning Rates for Streaming 3D Reconstruction}

\author{
Xinze Li\textsuperscript{1},
Yiyuan Wang\textsuperscript{1,2},
Pengxu Chen\textsuperscript{3},
Weifeng Su\textsuperscript{1,4},
Weisi Lin\textsuperscript{5},
Wentao Cheng\textsuperscript{1}\thanks{Corresponding author: Wentao Cheng. Email: wentaocheng@bnbu.edu.cn}
\\
\textsuperscript{1}Beijing Normal-Hong Kong Baptist University 
\\
\textsuperscript{2}Hong Kong Baptist University \\
\textsuperscript{3}Jilin University \\
\textsuperscript{4}Guangdong Provincial Key Laboratory of Interdisciplinary Research and Application for Data Science \\
\textsuperscript{5}Nanyang Technological University \\
}

\iclrfinalcopy 
\begin{document}
\maketitle

\begin{figure}[htbp]
    \centering
    \includegraphics[width=\linewidth]{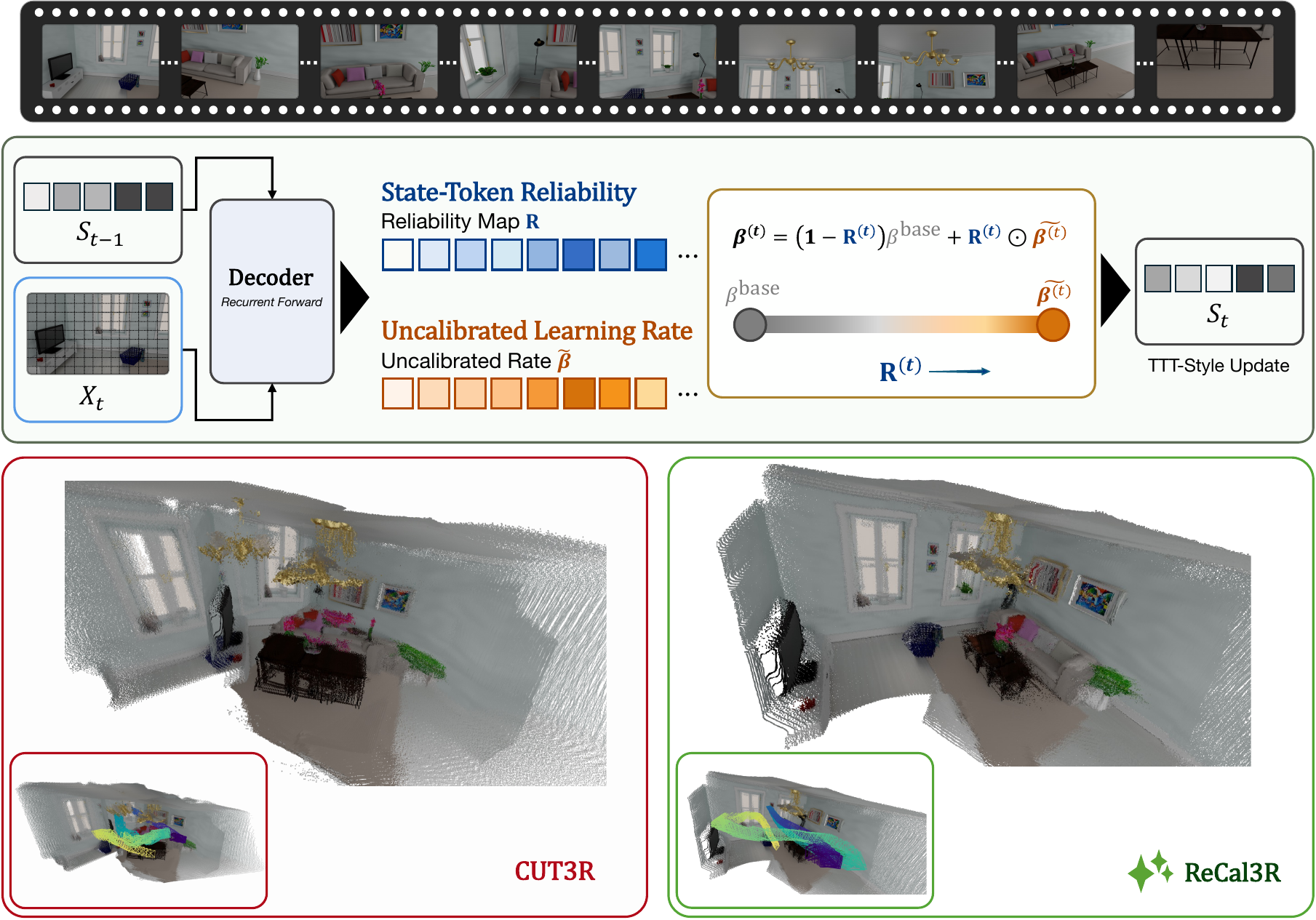}
    \caption{Given the current image and the recurrent scene state, ReCal3R derives a candidate learning rate and estimates state token reliability from intermediate signals produced by the recurrent forward pass. This prevents unreliable state tokens from receiving aggressive updates and leads to cleaner reconstructions than CUT3R over long image streams.}
    \label{fig:top_fig}
\end{figure}

\begin{abstract}
Streaming 3D reconstruction relies on a compact recurrent scene state to process long image streams in linear time and bounded memory. However, repeated updates can gradually corrupt this state, causing reliable historical information to be overwritten by noisy or ambiguous observations. We introduce ReCal3R, a reliability-calibrated learning rate method for recurrent 3D reconstruction. Instead of directly applying a candidate learning rate, our method estimates state token reliability from the maintained scene state and uses it to calibrate a candidate learning rate derived from token alignment, state reconstruction residual, and recent update pressure. The resulting token-wise learning rate interpolates between a conservative base rate and the candidate rate, suppressing aggressive updates on unreliable tokens while preserving adaptation to informative frames. Applied to CUT3R as a training-free calibration rule, ReCal3R reaches strong performance on long sequences in pose, depth, and reconstruction quality, including a 3.7$\times$ reduction in ATE, with comparable runtime and memory. Code is available at: \url{https://github.com/Powertony102/ReCal3R}

\end{abstract}

\section{Introduction}
\label{sec:intro}
Embodied agents, AR/VR systems, and mobile robots increasingly operate as continuous observers of the world: they consume image streams that are unbounded in length, arrive in real time, and must be integrated into a coherent geometric understanding on the fly. Recent feed-forward 3D geometry models such as DUSt3R~\citep{dust3r} and VGGT~\citep{vggt} have shown impressive geometry, pose, and correspondence prediction from RGB image sets. Yet they are primarily designed for finite view collections rather than online state maintenance. Streaming 3D reconstruction reframes the problem: a bounded scene representation must be continuously updated as new frames arrive, enabling linear time processing with bounded memory.

Recurrent reconstruction implements this formulation with a compact latent scene state that is updated and read out online. CUT3R~\citep{cut3r} is the canonical recurrent state model in this space, compressing the growing history into a bounded state. However, compact recurrence alone does not guarantee accuracy over long streams. As the input sequence grows, errors can accumulate inside the state, and reliable historical content may be overwritten by noisy or ambiguous observations. The central difficulty is therefore not only how to bound the representation, but how to update the bounded state reliably over time.

This limitation has motivated inference time modifications of recurrent state updates. TTT3R~\citep{ttt3r} casts the recurrent transition as Test-Time Training~\citep{ttt}, treating the scene state as fast weights and deriving a closed-form per-token learning rate from the alignment between the maintained state and incoming observations. Recent and concurrent methods further adjust the update policy using temporal and spatial adaptive signals, or by selectively updating memory components~\citep{ttsa3r,memix}. These methods improve stability over long streams, but their update rules are still largely driven by the current observation and its relation to the maintained state. They do not explicitly measure whether the state tokens being updated remain reliable carriers of historical geometry.

This missing diagnosis becomes important over long streams. A state token that repeatedly absorbs weak or ambiguous evidence may drift into regions of the latent space that are no longer decoded reliably. A later frame can still appear compatible with this degraded token and induce a large update, further amplifying state corruption. Thus, a candidate learning rate should not be used as the final update strength by default; it should be calibrated according to the reliability of the state tokens that receive the update.

We introduce ReCal3R, a reliability-calibrated learning rate method for streaming 3D reconstruction. ReCal3R first forms a candidate learning rate that combines token alignment, state reconstruction residual, and recent update pressure, so that updates are encouraged by well matched and unexplained observations while being reduced on repeatedly updated tokens. It then estimates state token reliability from the maintained scene state and uses this reliability to calibrate the candidate rate. The final learning rate interpolates between a conservative base rate and the candidate rate according to state token reliability. Reliable tokens can still adapt to informative observations, while unreliable tokens are protected from overly aggressive updates. The method requires no training and can be directly applied to CUT3R without modifying the recurrent architecture.

On ScanNet~\citep{dai2017scannet} sequences of 1,000 frames, ReCal3R achieves strong performance on long streams across pose, depth, and reconstruction quality. It reduces ATE by $3.7\times$ over CUT3R at the longest evaluated sequence length, while maintaining comparable runtime and memory. Across depth and reconstruction benchmarks, ReCal3R preserves stable geometry where prior streaming baselines gradually degrade.

In summary, ReCal3R advances streaming 3D reconstruction in three aspects:
\begin{enumerate}
    \item We reveal a reliability gap in recurrent state writing: existing adaptive updates mainly follow the incoming observation, but do not explicitly assess whether the state token being updated remains a reliable carrier of historical geometry.
    \item We propose ReCal3R, a learning rate calibration method that first constructs a candidate rate from token alignment, state reconstruction residual, and recent update pressure, and then calibrates it using state token reliability. The mechanism is closed-form, requires no training, and can be applied on top of CUT3R.
    \item We show strong performance on long streams in pose, depth, and reconstruction quality, while preserving the linear time and bounded memory formulation of recurrent reconstruction.
\end{enumerate}

\section{Related Work}
\label{sec:related-work}

\subsection{Generalizable 3D Reconstruction}

Classical Structure-from-Motion (SfM) pipelines~\cite{schonberger2016structure,pan2024global,zhu2018very,ni2007out} recover geometry through iterative correspondence estimation and bundle adjustment, but their optimization and matching costs limit real time and long stream perception. Recent 3D reconstruction foundation models instead predict dense geometry and cameras directly from RGB images in a feed-forward manner. One major line follows the shared coordinate pointmap paradigm introduced by DUSt3R~\cite{dust3r}, which directly regresses dense 3D structure from image pairs without requiring known camera parameters. This formulation has been extended by MASt3R~\cite{mast3r}, which grounds dense matching in metric 3D space. Follow up works such as MonST3R~\cite{monst3r} adapt the pointmap formulation to dynamic scenes, while Fast3R~\cite{fast3r} improves multi-view scalability within a feed-forward pipeline.

A parallel line is represented by VGGT~\cite{vggt}, which departs from pairwise pointmap matching and performs global geometry reasoning over a finite set of input views. Despite their architectural differences, these feed-forward formulations are primarily designed for finite view collections rather than online state maintenance. Streaming reconstruction therefore requires additional mechanisms for selecting, caching, compressing, or recurrently updating observations over time.

\subsection{Streaming 3D Reconstruction with Persistent Memory}

To support online reconstruction, recent methods introduce explicit or implicit memory mechanisms across frames. Spann3R~\cite{spann3r} maintains a spatial memory, CUT3R~\cite{cut3r} uses a compact recurrent state, and Point3R~\cite{point3r} anchors memory to explicit 3D pointers for stronger long range recall. Another route adapts VGGT style reasoning to causal inference with key value caching or cache compression. Representative methods include STream3R~\cite{stream3r}, StreamVGGT~\cite{streamvggt}, and InfiniteVGGT~\cite{infinitevggt}, as well as recent long stream systems such as STAC~\cite{stac}, LongStream~\cite{longstream}, and HorizonStream~\cite{horizonstream}.

These methods reduce the cost of processing long streams either by caching attention histories or by folding observations into a compact state. The recurrent route avoids maintaining a full attention history, but makes the quality of each state update critical: aggressive updates can overwrite stable historical content, while conservative updates may miss newly observed geometry.

\subsection{Test-Time Training in Streaming 3D Reconstruction}

Test-Time Training (TTT) provides a useful view of online state adaptation. TTT layers~\cite{ttt} interpret recurrent hidden states as fast weights updated at inference time, offering a way to compress long contexts into a finite state. Test-Time Training Done Right~\cite{tttdoneright} further improves this formulation with larger update chunks, strengthening the efficiency and capacity of test-time state updates. This view is related to recurrent alternatives to full attention, including linear attention~\cite{transformers_are_rnns}, state space models such as Mamba~\cite{mamba}, and test-time memory mechanisms such as Titans~\cite{titans}.

Within 3D reconstruction, Test3R~\cite{test3r} performs inference time geometric adaptation by optimizing cross view consistency at test time. More closely related to streaming reconstruction, TTT3R~\cite{ttt3r} treats the recurrent state in CUT3R as fast weights and modulates state updates according to the alignment confidence between the current observation and the maintained state. Recent and concurrent methods further modify the state update policy. TTSA3R~\cite{ttsa3r} uses temporal state evolution and spatial observation quality to regulate updates, while MeMix~\cite{memix} reduces unnecessary writing by selectively updating memory components. These methods improve stability over long streams, but their update rules are still mainly driven by the current observation, temporal consistency, or memory selection. They do not explicitly measure whether the state tokens being modified have themselves become unreliable after repeated updates.

A complementary TTT based direction scales feed-forward reconstruction rather than modifying recurrent state updates. VGG-T3~\cite{vggt3} replaces the variable length key value representation in VGGT style global attention with fixed size MLP fast weights. ZipMap~\cite{zipmap} and tttLRM~\cite{tttlrm} use TTT layers to aggregate image observations into compact scene states. These methods build dedicated large scale reconstruction models for efficient global context compression, while ReCal3R targets an existing recurrent reconstruction model and modifies only its update rule.

\section{Method}
\label{sec:method}

\subsection{Preliminaries: Stateful Recurrent Update}
\label{sec:preliminaries}

\paragraph{Problem Setup.}
Streaming 3D reconstruction processes an image stream $\{\mathbf{I}^{(t)}\}_{t=1}^{T}$ on-the-fly, distilling scene context into a persistent latent state. At each step $t$, an encoder maps the input frame $\mathbf{I}^{(t)}$ into image tokens $\mathbf{X}^{(t)} \in \mathbb{R}^{K \times C}$, while the model carries a compact state $\mathbf{S}^{(t-1)} \in \mathbb{R}^{M \times C}$ summarizing all past frames. Each row of the compact state acts as a state token, which serves as a persistent latent slot for historical scene content. We denote by $\mathbf{A}^{(t)} \in \mathbb{R}^{M \times K}$ the pre-softmax cross-attention logits between state tokens and image tokens, which characterize their alignment.

\paragraph{Uniform Recurrent Update.}
The vanilla recurrent framework of CUT3R~\cite{cut3r} writes the current frame into the state with uniform intensity:
\begin{equation}
    \mathbf{S}^{(t)} = \mathbf{S}^{(t-1)} + \mathrm{softmax}_{j}\!\left(\mathbf{A}^{(t)}\right)\mathbf{V}_{\mathbf{X}^{(t)}},
    \label{eq:cut3r-update}
\end{equation}
where $\mathbf{V}_{\mathbf{X}^{(t)}}$ is the value projection of the current image tokens and $\mathrm{softmax}_{j}$ normalizes along the image-token dimension. This update treats the write strength as uniform across state tokens, without explicitly accounting for whether the current frame is redundant or whether a state token remains reliable after repeated updates. As a result, useful historical content may be gradually overwritten in long streams.

\paragraph{Test-Time-Training View.}
TTT3R~\cite{ttt3r} reinterprets the recurrent transition as test-time training, in which the state plays the role of fast weights adapted at inference:
\begin{equation}
    \mathbf{S}^{(t)} = \mathbf{S}^{(t-1)} - \boldsymbol{\beta}^{(t)} \odot \boldsymbol{\nabla}\!\left(\mathbf{S}^{(t-1)},\mathbf{X}^{(t)}\right),
    \label{eq:ttt-update}
\end{equation}
where $\boldsymbol{\nabla}(\cdot,\cdot)$ is the update direction induced by the interaction between the state and the current frame, and $\boldsymbol{\beta}^{(t)}\in[0,1]^{M\times 1}$ is a per-token learning rate. By modulating the step size, Eq.~(\ref{eq:ttt-update}) generalizes the uniform addition of Eq.~(\ref{eq:cut3r-update}) and exposes a natural interface for controlling how aggressively each state token is overwritten.

\paragraph{Motivation.}
TTT3R derives the learning rate from the alignment between state tokens and image tokens. However, alignment alone is not a complete measure of update utility: \emph{a rich new frame may be weakly aligned simply because the current state only contains textureless anchors.} This suggests two questions beyond raw alignment: how much new content the current frame brings to the state, and whether the corresponding state token is reliable enough to support a strong update. ReCal3R answers these questions separately by estimating an uncalibrated learning rate and calibrating it with state token reliability.

\subsection{Overview: State Reliability for Calibrated Updating}
\label{sec:overview}

The motivation above suggests that alignment alone is insufficient for controlling long stream state updates. In ReCal3R, we therefore place state token reliability at the center of the update mechanism. Before directly following an uncalibrated learning rate, the model first estimates whether the corresponding state tokens remain reliable under long-term recurrent updates. The resulting reliability scores calibrate the final learning rate vector by interpolating between a conservative baseline and the uncalibrated rate.

\paragraph{Reliability-Calibrated Learning Rate.}
We define the final learning rate vector as
\begin{equation}
    \boldsymbol{\beta}^{(t)}
    =
    \left(\mathbf{1}-\boldsymbol{\mathcal{R}}^{(t)}\right)\beta^{\mathrm{base}}
    +
    \boldsymbol{\mathcal{R}}^{(t)} \odot \tilde{\boldsymbol{\beta}}^{(t)},
    \label{eq:calibrated-rate}
\end{equation}
where $\beta^{\mathrm{base}}$ is a small conservative rate, $\tilde{\boldsymbol{\beta}}^{(t)}$ is the uncalibrated learning rate vector, $\boldsymbol{\mathcal{R}}^{(t)}$ collects the reliability scores of all state tokens, $\mathbf{1}$ is an all-one vector, and $\odot$ denotes element-wise multiplication. Eq.~(\ref{eq:calibrated-rate}) performs an element-wise interpolation between the conservative baseline and the uncalibrated rate. Reliable state tokens allow the final rate to approach the uncalibrated rate, while unreliable tokens keep it close to the conservative baseline. Substituting $\boldsymbol{\beta}^{(t)}$ into Eq.~(\ref{eq:ttt-update}) yields the ReCal3R update.

\paragraph{State Token Reliability (\S\ref{sec:state-token-reliability}).}
State token reliability estimates how safely a state token can support a strong update. It considers both the token's accumulated deviation from its learned initialization and the concentration of its attention on the current frame. A large deviation suggests that the token has been repeatedly modified, while diffuse attention indicates an ambiguous relation to the current frame. A low reliability score therefore keeps the final learning rate close to the conservative baseline, instead of directly following an aggressive uncalibrated rate.

\paragraph{Uncalibrated Learning Rate (\S\ref{sec:uncalibrated-rate}).}
The uncalibrated learning rate estimates the candidate update strength before reliability calibration. It combines the alignment gate inherited from TTT3R, a state reconstruction residual score, and recent update pressure. The residual score increases the rate when the current frame contains content not well explained by the accumulated state, while recent update pressure suppresses tokens that have been activated repeatedly. This gives a candidate learning rate vector that reflects alignment, unexplained frame content, and update saturation, but is not applied directly. Together, state token reliability and the uncalibrated learning rate separate estimating a candidate update strength from deciding how much it should influence the final learning rate.

\subsection{State Token Reliability}
\label{sec:state-token-reliability}

Long stream reconstruction repeatedly writes new information into the same compact state. As a result, a state token may become unreliable for two different reasons. First, it may have accumulated large changes from its learned initialization, indicating that the token has been heavily modified by previous updates. Second, even if the token itself remains stable, its relation to the current frame may be ambiguous when its attention is diffuse over many image tokens. ReCal3R therefore estimates state token reliability from these two aspects: accumulated state deviation and attention entropy. The resulting reliability score does not define a new update direction; it only calibrates how much the uncalibrated learning rate should influence the final learning rate in Eq.~(\ref{eq:calibrated-rate}).

\paragraph{State Deviation and Attention Entropy.}
For each state token $m$, we first measure its accumulated deviation from the learned initialization as
$\Delta_m^{(t)}=\|\mathbf{S}_m^{(t-1)}-\mathbf{S}_m^{(0)}\|_2$.
We min-max normalize this deviation over all state tokens at the current step, yielding
$d_m^{(t)}\in[0,1]$. We also measure the ambiguity of the token-frame relation using normalized attention entropy. Let
\begin{equation}
    \mathbf{a}_m^{(t)}
    =
    \mathrm{softmax}\!\left(\mathbf{A}_{m,:}^{(t)}\right),
    \qquad
    e_m^{(t)}
    =
    \bar{\mathcal{H}}\!\left(\mathbf{a}_m^{(t)}\right),
    \label{eq:attention-entropy}
\end{equation}
where $\mathbf{a}_m^{(t)}$ is the attention distribution from state token $m$ to the current image tokens, and $\bar{\mathcal{H}}(\cdot)$ denotes Shannon entropy normalized by its maximum possible value. Thus $e_m^{(t)}\in[0,1]$, with a large value indicating diffuse attention and an ambiguous relation to the current frame.

In the following reliability pooling, we use the complements $1-d_m^{(t)}$ and $1-e_m^{(t)}$ as positive cues, corresponding to state stability and attention concentration, respectively.

\paragraph{Reliability from Cue Agreement.}
For clarity, we first describe the pooling for a single state token at one time step, and omit the token and time indices. Given normalized deviation $d$ and normalized entropy $e$, we combine state stability $1-d$ and attention concentration $1-e$ through an agreement-based pooling function:
\begin{equation}
    \rho
    =
    \frac{(1-d)(1-e)}
    {(1-d)(1-e) + de}.
    \label{eq:pooling}
\end{equation}
Applying Eq.~(\ref{eq:pooling}) to each state token at each time step gives the intermediate reliability estimate $\rho_m^{(t)}$.

Eq.~(\ref{eq:pooling}) assigns high reliability only when the state token is both stable and clearly associated with the current frame. It assigns low reliability when both cues indicate unreliability. When the two cues disagree, the estimate remains in an ambiguous range instead of making an overconfident decision. This behavior is useful because sharp attention alone should not make a heavily modified token reliable, and a stable token should not receive a strong learning rate when its attention to the current frame is diffuse. 

\paragraph{Confidence Calibration.}
The intermediate estimate $\rho$ is still unreliable when it lies near the ambiguous value $0.5$. We use the distance from this maximally ambiguous point to measure how decisive the estimate is:
\begin{equation}
    \mathcal{R}
    =
    \rho(2\rho - 1)^2.
    \label{eq:final-reliability}
\end{equation}
The factor $(2\rho - 1)^2$ is close to zero when $\rho$ is ambiguous and close to one when $\rho$ is decisive. Multiplying it by $\rho$ makes the final reliability high only when the state token is both reliable and confidently diagnosed as such. Applying Eq.~(\ref{eq:final-reliability}) to each state token at each time step gives the reliability vector $\boldsymbol{\mathcal{R}}^{(t)}$ used in Eq.~(\ref{eq:calibrated-rate}). Beyond these intuitive constructions, both the agreement-based pooling in Eq.~(\ref{eq:pooling}) and the confidence weighting in Eq.~(\ref{eq:final-reliability}) admit a probabilistic interpretation: under a single latent-variable model of state token reliability, $\rho$ coincides with the Bernoulli posterior over a binary reliability latent, while $\mathcal{R}$ is its confidence-weighted counterpart that down-weights ambiguous diagnoses. We defer this derivation, which further recasts the reliability-calibrated update in Eq.~(\ref{eq:calibrated-rate}) as a soft two-regime write policy, to Appendix~\ref{app:trust-derivation}.

\subsection{Uncalibrated Learning Rate}
\label{sec:uncalibrated-rate}

After estimating state token reliability, ReCal3R computes the learning rate vector before reliability calibration. This uncalibrated rate should be large only when three conditions are jointly satisfied: the current frame is well aligned with the state tokens, the accumulated state cannot fully explain the current frame, and the same state tokens have not been repeatedly updated in recent frames. We define
\begin{equation}
    \tilde{\boldsymbol{\beta}}^{(t)}
    =
    r^{(t)}\,\mathbf{g}^{(t)} \odot \phi\!\left(\mathbf{h}^{(t)}\right),
    \label{eq:uncalibrated-rate}
\end{equation}
where $\tilde{\boldsymbol{\beta}}^{(t)}$ is the uncalibrated learning rate vector, $\mathbf{g}^{(t)}$ is the token-wise alignment gate derived by TTT3R~\cite{ttt3r}, $r^{(t)}$ is a state reconstruction residual score shared by all tokens, $\mathbf{h}^{(t)}$ is a nonnegative update pressure vector over state tokens, and $\odot$ denotes element-wise multiplication. The function $\phi$ maps update pressure to an attenuation factor; we instantiate it element-wise as $\phi(x)=\exp(-x)$.

\paragraph{State Reconstruction Residual Score.}
The residual score measures how well the accumulated state explains the current frame. We reconstruct each projected image token $\tilde{\mathbf{x}}_k^{(t)}=(\mathbf{V}_{\mathbf{X}^{(t)}})_k$ from the state through the transposed attention relation, denote the reconstruction by $\hat{\mathbf{x}}_k^{(t)}$, and form the mean reconstruction residual
\begin{equation}
\epsilon^{(t)}
=
\frac{1}{K}\sum_{k=1}^{K}
\left\|
\tilde{\mathbf{x}}_k^{(t)}-\hat{\mathbf{x}}_k^{(t)}
\right\|_2^2.
\label{eq:state-residual}
\end{equation}
The residual score $r^{(t)}\in[0,1]$ is obtained by magnitude-normalizing $\epsilon^{(t)}$ and passing it through a sigmoid with a margin $\tau_r$ that absorbs irreducible reconstruction noise; the construction of $\hat{\mathbf{x}}_k^{(t)}$ and the exact mapping are deferred to Appendix~\ref{app:residual-score}. Thus $r^{(t)}$ grows when the current frame carries structure the accumulated state cannot predict, so a redundant but strongly aligned frame contributes little to the learning rate.

\paragraph{Recent Update Pressure.}
The update-pressure vector summarizes the recent write history as an exponential moving average of the per-token alignment gate, $h_m^{(t)} = \lambda h_m^{(t-1)} + (1-\lambda) g_m^{(t)}$, with decay $\lambda = 0.95$ (effective window $1/(1-\lambda)\approx 20$ frames). Through $\phi(x)=\exp(-x)$, recently saturated tokens receive a smaller uncalibrated rate, leaving reliability calibration to decide whether a token can safely follow it.

\paragraph{Multiplicative Form.}
The product in Eq.~(\ref{eq:uncalibrated-rate}) makes the uncalibrated learning rate depend on alignment, state reconstruction residual, and recent update pressure at the same time. Weak alignment lowers the token-specific rate, a small residual score suppresses updates for frames already explained by the state, and high update pressure suppresses tokens that have recently received frequent updates. Thus, $\tilde{\boldsymbol{\beta}}^{(t)}$ serves as a candidate learning rate vector, but it is not applied directly; the final learning rate is obtained after reliability calibration in Eq.~(\ref{eq:calibrated-rate}).

\section{Experiments}
We evaluate ReCal3R on three streaming 3D tasks: camera pose estimation (Section~\ref{sec:exp-camera-pose-estimation}), 3D reconstruction (Section~\ref{sec:exp-3d-reconstruction}), and video depth estimation (Section~\ref{sec:exp-video-depth-estimation}). These tasks evaluate complementary aspects of long stream reconstruction: global camera consistency, accumulated 3D geometry, and dense depth prediction over time.

\paragraph{Baselines.}
We compare ReCal3R with representative streaming reconstruction methods under the same online evaluation protocol. CUT3R~\cite{cut3r} serves as the compact recurrent baseline with fixed state updates. TTT3R~\cite{ttt3r} introduces a Test-Time Training view and modulates state updates using the alignment between the current observation and the maintained state. For a fair comparison, we adopt the TTT3R-based variant of MeMix~\cite{memix} as our baseline, as it employs the same token alignment strategy as our method; we refer to this variant simply as MeMix throughout the experiments. TTSA3R~\cite{ttsa3r} uses temporal state evolution and spatial observation quality to regulate state updates. ReCal3R differs from these methods by constructing a candidate learning rate and calibrating it with state token reliability estimated from the maintained scene state.

\paragraph{Evaluation Protocol.}
All methods are evaluated in streaming order without access to future frames. Unless otherwise stated, we use the official pretrained weights and follow the input resolution and preprocessing protocol of CUT3R. For each dataset, all methods are evaluated on the same valid sequences under the standard split, using identical temporal order and sampled frame indices. Runtime and peak GPU memory are measured on a single NVIDIA RTX PRO 6000 GPU with 96 GB of device memory. We use the standard metric for each task: ATE after Sim(3) alignment for camera pose estimation, accuracy, completeness, and normal consistency for 3D reconstruction, and Abs Rel together with $\delta < 1.25$ for video depth estimation.

\subsection{Camera Pose Estimation}
\label{sec:exp-camera-pose-estimation}

We evaluate camera pose accuracy on TUM-Dynamics~\cite{tum-d} and ScanNet~\cite{dai2017scannet}, following the evaluation protocol of prior 3D reconstruction methods~\cite{cut3r,mast3r,monst3r}. We report Absolute Trajectory Error (ATE) after Sim(3) alignment between the estimated and ground truth camera trajectories. To evaluate long stream robustness, we report results over increasing numbers of input frames.

\begin{figure}[htbp]
    \centering
    \includegraphics[width=1\linewidth]{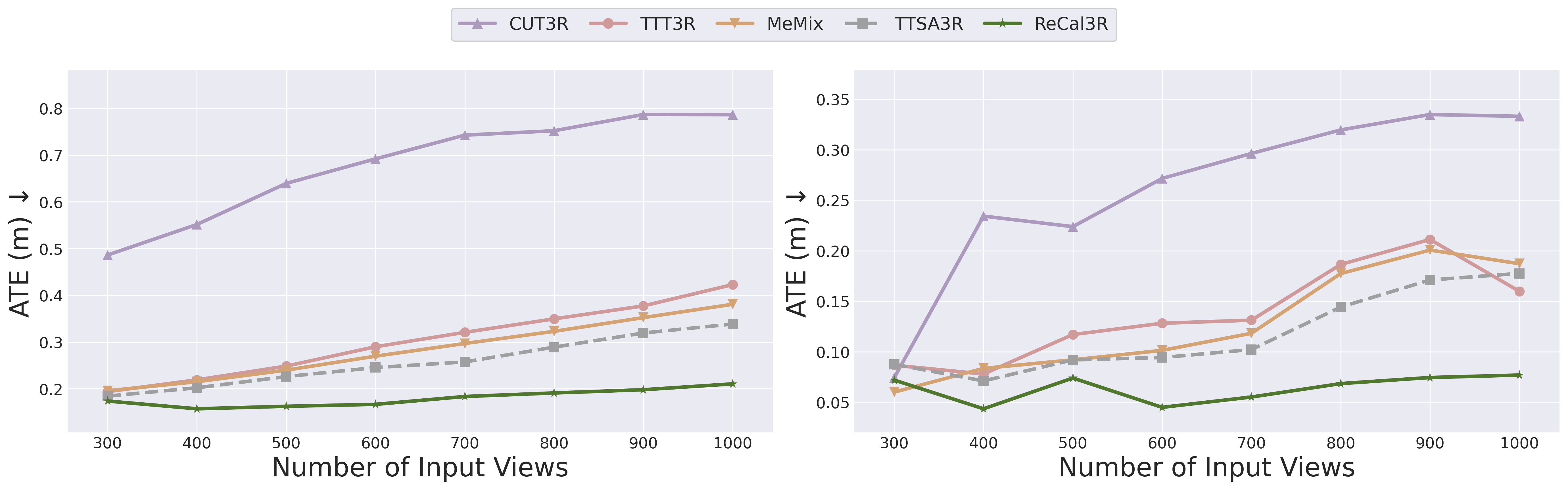}
    \caption{
    \textbf{Camera pose estimation.}
    Absolute Trajectory Error (ATE, $\downarrow$) on ScanNet~\cite{dai2017scannet} (left) and TUM-Dynamics~\cite{tum-d} (right) over increasing stream lengths. Camera trajectories are evaluated after Sim(3) alignment to the ground truth trajectories.
    }
    \label{fig:quant-relpose}
\end{figure}

Fig.~\ref{fig:quant-relpose} shows how pose accuracy changes as more frames are processed. CUT3R accumulates large trajectory error on both datasets, indicating that compact recurrence alone is not sufficient for stable pose estimation over long streams. TTT3R improves over CUT3R by adapting the state update strength from observation-state alignment, and MeMix and TTSA3R further reduce drift through selective memory updates or temporal and spatial update signals.

ReCal3R shows the clearest advantage in the long stream regime. On ScanNet, it maintains substantially lower ATE as the number of input frames increases, whereas the competing streaming baselines continue to accumulate drift. On TUM-Dynamics, some baselines are competitive at the shortest evaluated length, but their errors grow more rapidly as the stream becomes longer. In contrast, ReCal3R remains stable over longer sequences and achieves the lowest ATE in the later part of the evaluation. On $1{,}000$-frame ScanNet sequences, ReCal3R reduces ATE by $3.7\times$ compared with CUT3R while maintaining comparable runtime and memory. These results suggest that reliability calibration improves the stability of recurrent state updates, leading to more accurate camera trajectories over long image streams.

\subsection{3D Reconstruction}
\label{sec:exp-3d-reconstruction}

We evaluate 3D reconstruction on 7-Scenes~\cite{7scene} and NRGBD~\cite{nrgbd}. For both datasets, we use a keyframe sampling stride of 2 for evaluated stream lengths below 500 frames and a stride of 1 for evaluated stream lengths of 500 frames or longer. Following prior work, we compare the predicted pointmaps with the ground-truth point clouds and report accuracy, completeness, and normal consistency. Accuracy and completeness measure geometric distance in opposite directions, while normal consistency evaluates local surface quality.

\begin{table*}[htbp]
\centering
\caption{
\textbf{Multi view 3D reconstruction.}
Mean accuracy ($\downarrow$), completeness ($\downarrow$), and normal consistency ($\uparrow$) on 7-Scenes~\cite{7scene} and NRGBD~\cite{nrgbd} over increasing numbers of input frames. Best in \textbf{bold}, second best \underline{underlined}.
}
\label{tab:quant-mvrecon}
\scriptsize
\setlength{\tabcolsep}{2.5pt}

\begin{subtable}{\textwidth}
\centering
\caption{7-Scenes.}
\resizebox{\textwidth}{!}{%
\begin{tabular}{l ccc ccc ccc ccc ccc}
\toprule
\multirow{2}{*}{Method} & \multicolumn{3}{c}{300 Frames} & \multicolumn{3}{c}{400 Frames} & \multicolumn{3}{c}{500 Frames} & \multicolumn{3}{c}{600 Frames} & \multicolumn{3}{c}{700 Frames} \\
\cmidrule(lr){2-4}\cmidrule(lr){5-7}\cmidrule(lr){8-10}\cmidrule(lr){11-13}\cmidrule(lr){14-16}
& Acc. $\downarrow$ & Comp. $\downarrow$ & NC $\uparrow$ & Acc. $\downarrow$ & Comp. $\downarrow$ & NC $\uparrow$ & Acc. $\downarrow$ & Comp. $\downarrow$ & NC $\uparrow$ & Acc. $\downarrow$ & Comp. $\downarrow$ & NC $\uparrow$ & Acc. $\downarrow$ & Comp. $\downarrow$ & NC $\uparrow$ \\
\midrule
CUT3R & 0.144 & 0.077 & 0.557 & 0.177 & 0.114 & 0.549 & 0.200 & 0.099 & 0.544 & 0.197 & 0.144 & 0.540 & 0.210 & 0.130 & 0.539 \\
TTT3R & 0.044 & 0.028 & 0.602 & 0.057 & 0.031 & 0.595 & 0.070 & 0.038 & 0.586 & 0.085 & 0.042 & 0.573 & 0.099 & 0.051 & 0.570 \\
MeMix & 0.040 & 0.027 & 0.603 & 0.049 & 0.032 & 0.597 & 0.067 & 0.038 & 0.587 & 0.077 & 0.039 & 0.579 & 0.090 & 0.044 & 0.574 \\
TTSA3R & \underline{0.033} & \underline{0.025} & \underline{0.613} & \underline{0.041} & \textbf{0.024} & \underline{0.607} & \underline{0.051} & \underline{0.030} & \underline{0.600} & \underline{0.057} & \underline{0.031} & \underline{0.593} & \underline{0.069} & \underline{0.035} & \underline{0.588} \\
\textbf{ReCal3R} & \textbf{0.023} & \textbf{0.024} & \textbf{0.619} & \textbf{0.025} & \underline{0.026} & \textbf{0.618} & \textbf{0.029} & \textbf{0.026} & \textbf{0.614} & \textbf{0.031} & \textbf{0.026} & \textbf{0.610} & \textbf{0.034} & \textbf{0.028} & \textbf{0.608} \\
\bottomrule
\end{tabular}
}
\end{subtable}

\vspace{0.8em}

\begin{subtable}{\textwidth}
\centering
\caption{NRGBD.}
\resizebox{\textwidth}{!}{%
\begin{tabular}{l ccc ccc ccc ccc ccc}
\toprule
\multirow{2}{*}{Method} & \multicolumn{3}{c}{300 Frames} & \multicolumn{3}{c}{400 Frames} & \multicolumn{3}{c}{500 Frames} & \multicolumn{3}{c}{600 Frames} & \multicolumn{3}{c}{700 Frames} \\
\cmidrule(lr){2-4}\cmidrule(lr){5-7}\cmidrule(lr){8-10}\cmidrule(lr){11-13}\cmidrule(lr){14-16}
& Acc. $\downarrow$ & Comp. $\downarrow$ & NC $\uparrow$ & Acc. $\downarrow$ & Comp. $\downarrow$ & NC $\uparrow$ & Acc. $\downarrow$ & Comp. $\downarrow$ & NC $\uparrow$ & Acc. $\downarrow$ & Comp. $\downarrow$ & NC $\uparrow$ & Acc. $\downarrow$ & Comp. $\downarrow$ & NC $\uparrow$ \\
\midrule
CUT3R & 0.226 & 0.135 & 0.596 & 0.290 & 0.110 & 0.578 & 0.327 & 0.202 & 0.564 & 0.371 & 0.202 & 0.552 & 0.388 & 0.233 & 0.550 \\
TTT3R & 0.105 & 0.030 & 0.646 & 0.140 & 0.074 & 0.633 & 0.136 & 0.041 & 0.630 & 0.172 & 0.051 & 0.616 & 0.197 & 0.073 & 0.608 \\
MeMix & \textbf{0.063} & \underline{0.018} & \underline{0.670} & 0.134 & 0.066 & 0.634 & 0.109 & 0.034 & 0.634 & 0.135 & 0.041 & 0.623 & 0.160 & 0.055 & 0.614 \\
TTSA3R & 0.091 & 0.023 & 0.669 & \underline{0.106} & \underline{0.040} & \underline{0.665} & \underline{0.085} & \underline{0.025} & \underline{0.660} & \underline{0.106} & \underline{0.030} & \underline{0.651} & \underline{0.123} & \underline{0.043} & \underline{0.644} \\
\textbf{ReCal3R} & \underline{0.078} & \textbf{0.016} & \textbf{0.685} & \textbf{0.086} & \textbf{0.030} & \textbf{0.684} & \textbf{0.070} & \textbf{0.018} & \textbf{0.675} & \textbf{0.090} & \textbf{0.022} & \textbf{0.666} & \textbf{0.093} & \textbf{0.031} & \textbf{0.658} \\
\bottomrule
\end{tabular}
}
\end{subtable}
\end{table*}

\begin{figure}[htbp]
    \centering
    \includegraphics[width=\linewidth]{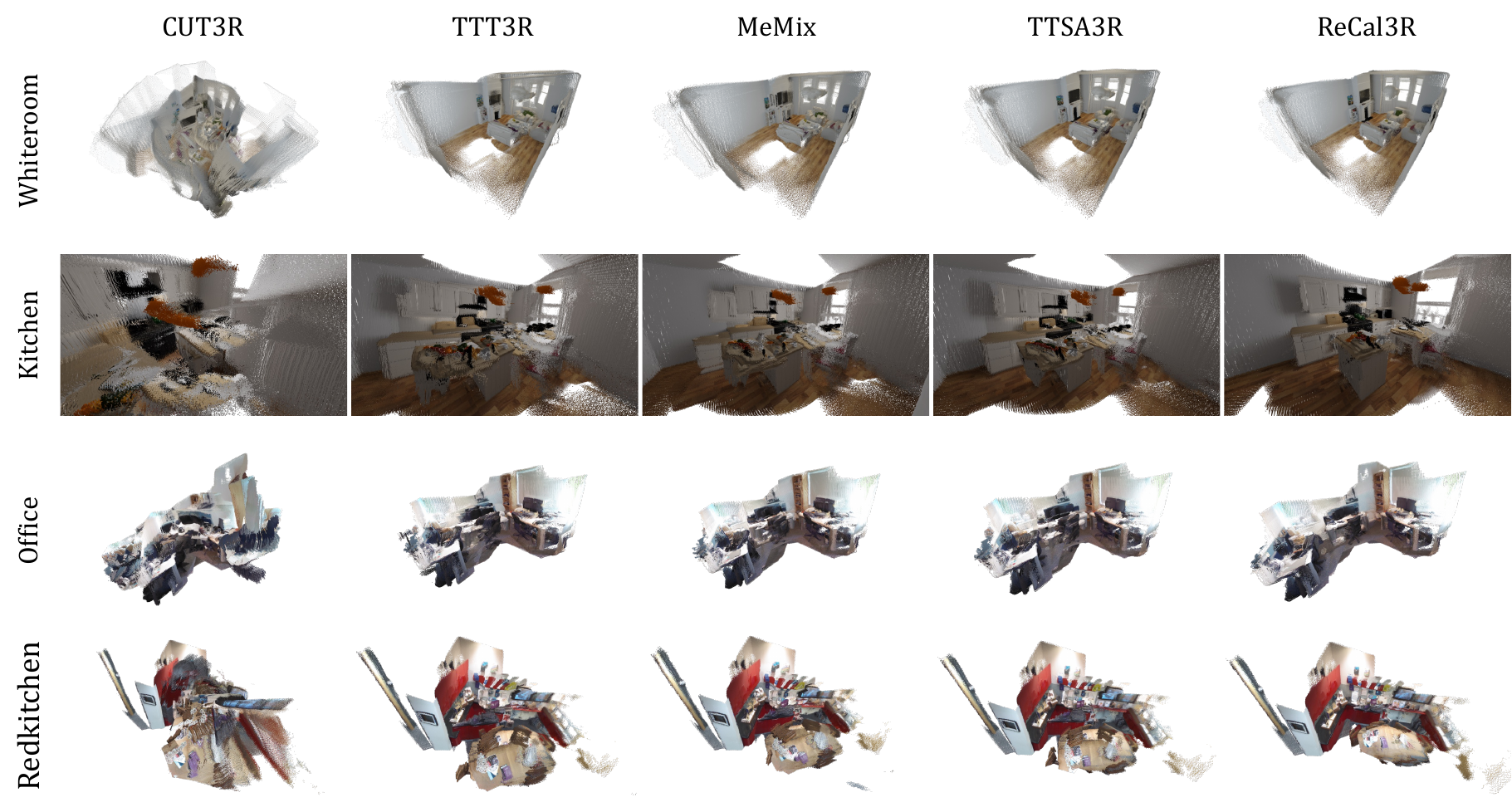}
    \caption{\textbf{Qualitative comparison for 3D reconstruction} on representative scenes from 7-Scenes and NRGBD. ReCal3R produces more coherent geometry under long-sequence reconstruction, with reduced structural distortion and better-preserved surface details.}
    \label{fig:qual-mvrecon}
\end{figure}

Tab.~\ref{tab:quant-mvrecon} reports quantitative reconstruction results under increasing stream lengths, and Fig.~\ref{fig:qual-mvrecon} provides qualitative comparisons. On 7-Scenes, ReCal3R achieves the strongest overall performance across all evaluated sequence lengths. Its advantage becomes more pronounced as more frames are accumulated, indicating that reliability calibration helps preserve the quality of the recurrent scene state during long stream reconstruction.

On NRGBD, MeMix attains the best accuracy at the shortest sequence length, but ReCal3R becomes stronger as the stream grows and achieves the best accuracy from 400 to 700 frames. It also obtains the best completeness and normal consistency across all evaluated lengths. These results suggest that ReCal3R is not merely improving a single metric, but more consistently preserves both geometric coverage and local surface quality over time.

The qualitative results in Fig.~\ref{fig:qual-mvrecon} show the same outcome. Baseline methods tend to accumulate distortions or incomplete surfaces as the recurrent state is updated over many frames. In contrast, ReCal3R produces more coherent scene geometry with fewer accumulated artifacts. This supports the role of state token reliability in calibrating candidate updates: unreliable state tokens are prevented from receiving overly aggressive updates, while reliable tokens can still incorporate newly observed geometry.

\subsection{Video Depth Estimation}
\label{sec:exp-video-depth-estimation}

We evaluate video depth estimation on Bonn~\cite{bonn} and TUM-Dynamics~\cite{tum-d}. We report Absolute Relative Error (Abs Rel) and the threshold accuracy $\delta < 1.25$. Following the standard protocol of each dataset, Bonn is evaluated with a single scale alignment for each sequence, while TUM-Dynamics is evaluated in metric scale without scale alignment. We report results over increasing stream lengths to evaluate temporal stability under longer inputs.

\begin{figure} [h]
    \centering
    \includegraphics[width=1\linewidth]{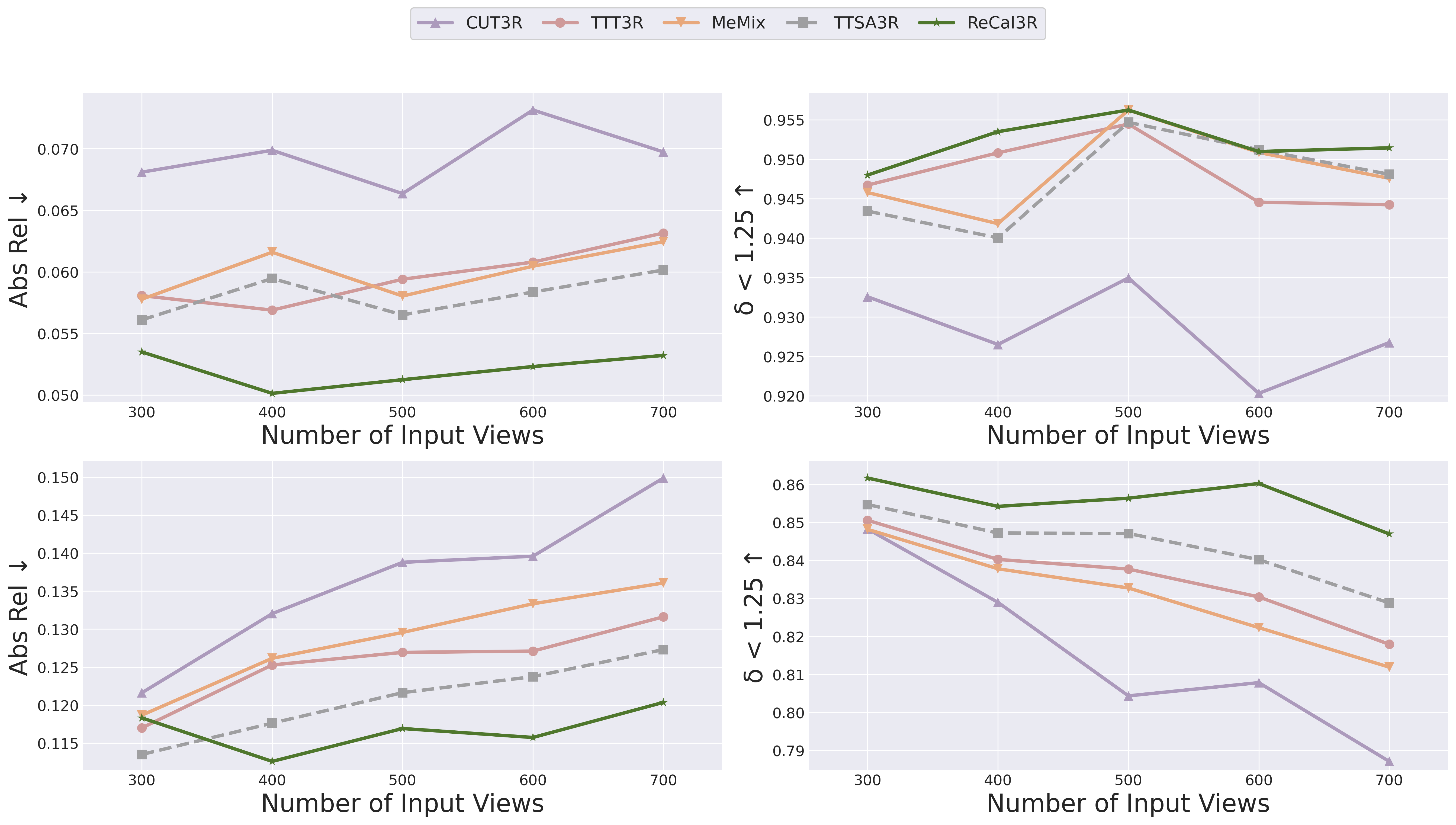}
    \caption{\textbf{Video depth estimation.} Abs Rel ($\downarrow$) and $\delta < 1.25$ ($\uparrow$) on Bonn~\cite{bonn} (top) and TUM-D~\cite{tum-d} (bottom) over increasing numbers of input views. Bonn is evaluated in scale mode with a single per-sequence scale alignment, while TUM-D is evaluated in metric mode without scale alignment.}
    \label{fig:quant-depth}
\end{figure}

Fig.~\ref{fig:quant-depth} compares depth prediction quality as the input stream becomes longer. CUT3R shows clear degradation on both datasets, especially on TUM-Dynamics, where Abs Rel increases and threshold accuracy drops as more frames are accumulated. TTT3R improves over CUT3R, and MeMix and TTSA3R further reduce errors in several settings by using selective memory updates or temporal and spatial update signals.

ReCal3R achieves the strongest overall performance across the two datasets and metrics, with the most visible advantage in longer streams. On Bonn, ReCal3R obtains the lowest Abs Rel across the evaluated lengths and remains competitive in threshold accuracy. On TUM-Dynamics, although some baselines are competitive at the shortest length, ReCal3R degrades more slowly as the stream grows and gives the best results in the longer sequence range. These trends suggest that reliability calibration helps preserve stable depth predictions during recurrent state updates, especially when accumulated state errors become more pronounced over time.

\subsection{Ablation Study}
\label{sec:exp-ablation}

Tab.~\ref{tab:ablation-pose} ablates the two main components of ReCal3R on 1,000-frame pose estimation: the uncalibrated learning rate and the reliability calibration. To isolate the contribution of the uncalibrated learning rate, we replace it with the original TTT3R rate while keeping reliability calibration. To isolate the contribution of reliability calibration, we remove the reliability term and directly use the uncalibrated learning rate as the final learning rate.

\begin{wraptable}{r}{0.48\linewidth}
    \centering
    \small
    \captionsetup{justification=raggedright,singlelinecheck=false,width=\linewidth}
    \setlength{\tabcolsep}{3pt}
    \caption{\textbf{State update ablation.} ATE ($\downarrow$) on 1,000-frame ScanNet and TUM-Dynamics sequences.}
    \label{tab:ablation-pose}
    \begin{tabularx}{\linewidth}{
        @{}
        >{\raggedright\arraybackslash}p{0.52\linewidth}
        >{\centering\arraybackslash}X
        >{\centering\arraybackslash}X
        @{}
    }
        \toprule
        \multirow{2}{*}{Method} & \multicolumn{2}{c}{Dataset} \\
        \cmidrule(lr){2-3}
        & ScanNet & TUM-D \\
        \midrule
        CUT3R & 0.7865 & 0.3330 \\
        TTT3R & 0.4230 & 0.1599 \\
        ReCal3R w/ TTT3R rate & 0.2431 & 0.1152 \\
        ReCal3R w/o reliability & 0.2502 & 0.1169 \\
        \textbf{ReCal3R} & \textbf{0.2106} & \textbf{0.0769} \\
        \bottomrule
    \end{tabularx}
\end{wraptable}

The variant with the TTT3R rate removes our uncalibrated learning rate construction and uses the original TTT3R rate as the candidate rate. Its improvement over TTT3R shows that reliability calibration is useful even when the candidate rate is inherited from an existing alignment-based rule.

The variant without reliability directly applies our uncalibrated learning rate, which combines token alignment, state reconstruction residual, and recent update pressure. It also improves over TTT3R, indicating that the proposed rate provides a stronger update signal than alignment alone. Combining this rate with reliability calibration gives the best performance, reducing ATE to 0.2106 on ScanNet and 0.0769 on TUM-Dynamics. These results show that ReCal3R benefits from both a more informative candidate rate and a reliability calibration step before state updating. We note that ReCal3R does not dominate every entry: on relatively short streams it is comparable to its variants, and its default setting trades a slightly lower long stream ATE for stable cross-length behavior, reflecting a balance between long-term stability and frame-level adaptivity.

The appendix further extends this decomposition to the other tasks and internal factors. Appendix~\ref{sec:app-pillar-ablation} evaluates the two variants on video depth estimation and 3D reconstruction. Appendix~\ref{sec:app-diagnostic-cues} separates the diagnostic cues used for state token reliability, while Appendices~\ref{sec:app-beta-ablation} and~\ref{sec:app-beta-base} analyze the effects of state reconstruction residual, recent update pressure, and the conservative base rate.

\section{Conclusion}

We presented ReCal3R, a reliability calibrated learning rate method for streaming 3D reconstruction. ReCal3R derives a candidate learning rate from token alignment, state reconstruction residual, and recent update pressure, and calibrates it using state token reliability estimated from the maintained scene state. This separates the update strength suggested by the incoming observation from the reliability of the state token that receives the update. The resulting closed form and training free rule improves long stream stability across pose estimation, video depth estimation, and 3D reconstruction while preserving the runtime and memory profile of CUT3R. One limitation is that conservative updates can occasionally reduce fine frame-level details, even though they improve the overall consistency of long stream reconstruction. Future work could extend reliability calibrated updates to other forms of streaming 3D models beyond compact recurrent states.

\bibliographystyle{iclr2026_conference}
\bibliography{iclr2026_conference}

\clearpage
\clearpage
\newpage
\appendix

\addcontentsline{toc}{section}{Appendix}
\renewcommand \thepart{}
\renewcommand \partname{}
\part{Appendix}
\parttoc

\section{Derivation of the Reliability Calibration}
\label{app:trust-derivation}

This appendix derives the pooled reliability $\rho$ (Eq.~\ref{eq:pooling})
and the calibrated reliability $\mathcal{R}$ (Eq.~\ref{eq:final-reliability}) of Sec.~\ref{sec:state-token-reliability}
from a single latent-variable model. Following the per-token presentation in the main text,
we consider one state token at one time step and omit the token and time indices throughout
the derivation. This construction offers a probabilistic interpretation of the two
quantities used in the main text: (i) under the product-likelihood model, the pooled
reliability $\rho$ coincides with a Bernoulli posterior, so its variance can be read as a
second-order statistic of the same belief; and (ii) the reliability-calibrated update of
Eq.~\ref{eq:calibrated-rate} can be viewed as the decision rule associated with this model.

\subsection{A Binary Reliability Latent}
\label{app:latent}

For the state token under consideration, we introduce a binary latent variable
\begin{equation}
  z \in \{0,1\},
\end{equation}
where $z=1$ asserts that the receiving token is a dependable carrier of
historical geometry and may safely absorb the evidence-driven write, while $z=0$
asserts that it should not. We stress that $z$ is an inference device only: it is
never thresholded into a hard write/freeze decision. The update direction and its
nominal magnitude $\tilde\beta$ remain governed by the attention mechanism;
$z$ serves solely to produce a calibrated soft reliability used to interpolate the
rate.

Absent any token-specific evidence we have no reason to presume reliability either way, so
we place a non-informative Bernoulli prior
\begin{equation}
  p\!\left(z=1\right)=\pi_0=\tfrac{1}{2}.
  \label{eq:prior}
\end{equation}
All token-dependent information enters exclusively through the likelihoods below, which
keeps the prior from acting as a hidden hyperparameter.

\subsection{Two Diagnostic Likelihoods}
\label{app:likelihood}

Reliability is assessed from two normalized cues: the state-side
deviation $d\in[0,1]$ (departure of the token from its learned initialization) and the
observation-side attention entropy $e\in[0,1]$ of the current
state-to-frame attention. Both are constructed so that a small value indicates a
reliable token. We therefore adopt the minimal Bernoulli-consistent forms, which are
monotone in this sense and symmetric between the two hypotheses,
\begin{equation}
  \begin{aligned}
    p\!\left(d \mid z\right)
    &\propto (1-d)^z d^{\,1-z}, \\
    p\!\left(e \mid z\right)
    &\propto (1-e)^z e^{\,1-z}.
  \end{aligned}
\end{equation}
A linear likelihood is a simple choice consistent with these two requirements that
introduces no additional shape hyperparameter; we adopt it for this reason, not because it
is uniquely determined.

\paragraph{Factorized likelihood.}
We model the joint likelihood as a product,
\begin{equation}
  p\!\left(d,e \mid z\right)
  = p\!\left(d \mid z\right)\,p\!\left(e \mid z\right).
  \label{eq:cond-indep}
\end{equation}
The two cues are read from different sources: $d$ from the token's accumulated deviation
from its learned initialization, and $e$ from the entropy of its current attention
distribution. This distinction motivates the product form and yields the closed-form
posterior below. The resulting rule is
conjunctive: $\rho\!\to\!1$ requires both cues to be healthy, so when they disagree the
gate defers to the more conservative one.

\subsection{Variational Posterior and Its Closed Form}
\label{app:elbo}

We approximate the intractable posterior $p\!\left(z \mid d,e\right)$ with a variational
Bernoulli $q(z)=\mathrm{Bernoulli}(\rho)$ and maximize the ELBO. Writing
$\ell_1=(1-d)(1-e)$ and $\ell_0=de$ for the unnormalized joint likelihoods under the two hypotheses,
\begin{equation}
  \mathcal{L}(\rho)
  = \mathbb{E}_{q}\!\left[\log p(d,e,z)\right]
    - \mathbb{E}_{q}\!\left[\log q(z)\right]
  = \rho \log\frac{\pi_0\,\ell_1}{\rho}
    + (1-\rho)\log\frac{(1-\pi_0)\,\ell_0}{1-\rho}.
  \label{eq:elbo}
\end{equation}
Differentiating and setting $\partial\mathcal{L}/\partial\rho=0$,
\begin{equation}
  \log\frac{\pi_0\,\ell_1}{\rho}-\log\frac{(1-\pi_0)\,\ell_0}{1-\rho}=0
  \;\Longrightarrow\;
  \frac{\pi_0\,\ell_1}{\rho}=\frac{(1-\pi_0)\,\ell_0}{1-\rho},
\end{equation}
which yields the closed-form optimum
\begin{equation}
  \rho^\star=\frac{\pi_0\,\ell_1}{\pi_0\,\ell_1+(1-\pi_0)\,\ell_0}.
\end{equation}
Substituting the non-informative prior $\pi_0=\tfrac12$ from Eq.~\ref{eq:prior} cancels the
prior factors and recovers the reliability of the main text,
\begin{equation}
  \rho
  = \frac{(1-d)(1-e)}
         {(1-d)(1-e)+de}.
  \label{eq:reliability-derived}
\end{equation}
This recovers the product opinion pool of the two cues: under the model of
Secs.~\ref{app:latent} and~\ref{app:likelihood}, $\rho$ coincides with the posterior
$p(z=1 \mid d,e)$. We use this correspondence to motivate the form of
Eq.~\ref{eq:pooling}, rather than to claim optimality among all possible fusion rules.

\paragraph{Posterior mean.} Since $z$ is binary, the Bernoulli family already contains the
exact posterior, so the optimum $\rho$ equals $p(z=1 \mid d,e)$ under this model:
\begin{equation}
  \rho=p\!\left(z=1 \mid d,e\right).
\end{equation}
We therefore treat $\rho$ as a posterior mean and its variance as the corresponding
second-order statistic in what follows.

\subsection{From Pooled Reliability to Calibrated Reliability}
\label{app:trust}

The posterior $q^\star=\mathrm{Bernoulli}(\rho)$ supplies two complementary
statistics: its mean records \emph{which} hypothesis is favored, while its variance records
\emph{how decisively}. The variance is
\begin{equation}
  \mathrm{Var}_{q^\star}[z]=\rho(1-\rho),
\end{equation}
which attains its maximum $\tfrac14$ at $\rho=\tfrac12$. Normalizing by this maximum
gives a unit-range ambiguity score
\begin{equation}
  v=\frac{\rho(1-\rho)}{1/4}=4\rho(1-\rho)\in[0,1],
\end{equation}
with $v\to1$ at maximal ambiguity and $v\to0$ when the posterior is
decisive. The corresponding confidence is its complement,
\begin{equation}
  c=1-v=1-4\rho(1-\rho)=(2\rho-1)^2,
  \label{eq:confidence-derived}
\end{equation}
recovering the confidence factor $(2\rho-1)^2$ in Eq.~\ref{eq:final-reliability}. Weighting the reliability by its own
confidence defines the calibrated reliability,
\begin{equation}
  \mathcal{R}=\rho(2\rho-1)^2.
  \label{eq:trust-derived}
\end{equation}
Identical to Eq.~\ref{eq:final-reliability}. Intuitively, $\rho$ chooses a direction and $c$ reports how
sure we are of it; a token earns an aggressive write only when it is both judged reliable
and judged so confidently. Near $\rho=\tfrac12$ the direction is essentially
a coin flip, $c\to0$ drives $\mathcal{R}\to0$, and the rate retreats to the
conservative baseline.

\subsection{The Reliability-Calibrated Update as a Two-Regime Policy}
\label{app:update}

Finally we connect $\mathcal{R}$ to the learning rate vector. For each token, the latent
suggests a soft two-regime reading: use the uncalibrated rate when the token is reliable
($z=1$), and the conservative rate $\beta^{\mathrm{base}}$ otherwise ($z=0$). Replacing the raw
posterior mean $\rho$ with the confidence-weighted reliability $\mathcal{R}$ routes
ambiguous diagnoses to the fallback rather than to a near-even blend. Applying this policy
element-wise to all state tokens gives exactly the vector interpolation of the main text,
\begin{equation}
    \boldsymbol{\beta}^{(t)}
    =
    \left(\mathbf{1}-\boldsymbol{\mathcal{R}}^{(t)}\right)\beta^{\mathrm{base}}
    +
    \boldsymbol{\mathcal{R}}^{(t)} \odot \tilde{\boldsymbol{\beta}}^{(t)},
\end{equation}
which is Eq.~\ref{eq:calibrated-rate}.

\subsection{State Reconstruction Residual Score \texorpdfstring{$r^{(t)}$}{r}}
\label{app:residual-score}

We detail the state reconstruction residual score deferred from Sec.~\ref{sec:uncalibrated-rate}. Let
$\tilde{\mathbf{X}}^{(t)}\equiv\mathbf{V}_{\mathbf{X}^{(t)}}$ denote the value projection of
the current image tokens $\mathbf{X}^{(t)}$ introduced in Sec.~\ref{sec:preliminaries}.
Analogously, $\mathbf{V}_{\mathbf{S}^{(t-1)}}$ denotes the value projection of the current
state. We reconstruct the projected image tokens by attending to this state-side value
projection through the transposed attention matrix,
\begin{equation}
  \hat{\mathbf{X}}^{(t)}
  =\mathrm{softmax}_m\!\left(\mathbf{A}^{(t)\top}\right)\mathbf{V}_{\mathbf{S}^{(t-1)}}.
\end{equation}
Both are already available from the standard recurrent forward pass, so computing $r^{(t)}$
requires no extra decoder call. The score is the sigmoid of the magnitude-normalized residual,
offset by a margin $\tau_r$ that absorbs irreducible reconstruction noise,
\begin{equation}
  r^{(t)}=\sigma\!\left(
    \frac{\tfrac{1}{K}\sum_{j=1}^{K}\bigl\lVert \tilde{\mathbf{X}}^{(t)}_j-\hat{\mathbf{X}}^{(t)}_j\bigr\rVert_2^2}
         {\tfrac{1}{K}\sum_{j=1}^{K}\bigl\lVert \tilde{\mathbf{X}}^{(t)}_j\bigr\rVert_2^2+\varepsilon}
    -\tau_r\right).
\end{equation}
When the frame carries structure the accumulated state cannot predict, the residual grows
and $r^{(t)}\!\to\!1$; for redundant frames the residual is small and $r^{(t)}$ attenuates,
so a redundant but strongly aligned frame contributes little to the write.

\section{Ablation Study}
\subsection{Decoupling Reliability and the Uncalibrated Rate}
\label{sec:app-pillar-ablation}

We disentangle the two controls behind ReCal3R's state update to verify that each contributes independently. The \emph{w/ TTT3R rate} variant replaces our uncalibrated rate with TTT3R's bare alignment rate while retaining reliability calibration, so any gain isolates the state-side \emph{whether} decision. The \emph{w/o reliability} variant keeps our uncalibrated rate, which combines alignment, state reconstruction residual, and update pressure, but disables reliability calibration, thereby isolating the observation-side \emph{how much} enrichment. Both variants outperform TTT3R across all three tasks. On 1000-frame ScanNet, ATE falls from 0.423 to 0.243 for \emph{w/ TTT3R rate} and to 0.250 for \emph{w/o reliability}; on 700-frame 7-Scenes, accuracy drops from 0.099 to 0.043 and 0.050, and normal consistency rises from 0.570 to 0.603 and 0.595. Notably, \emph{w/ TTT3R rate} differs from TTT3R only through reliability calibration, yet it already yields a large gain, showing that the reliability signal is not a disguised global learning rate reduction but a token-specific estimate. Conversely, \emph{w/o reliability} confirms that residual- and pressure-aware modulation suppresses redundant or saturated writes even when the receiving state is assumed reliable. The full model then achieves the strongest overall trade-off, including the best long stream ATE on ScanNet (0.211 at 1000 frames), the best relative depth error on Bonn (0.053 at 700 frames), the best reconstruction accuracy on 7-Scenes (0.034 at 700 frames), and the lowest rotational error on 300-frame ScanNet (RPE$_{\mathrm{rot}} = 0.901$). These results indicate that the two controls address complementary failure modes: reliability calibration handles representational deviation on the state side, while the uncalibrated rate handles observation-side redundancy and saturation. We therefore find that stable long stream state writing requires answering both \emph{whether} to write and \emph{how much} to write. We note a stability–plasticity trade-off: by enforcing conservative long stream writes, ReCal3R markedly improves global drift (ATE) while trading off a small amount of per-frame local sharpness (RPE), a pattern shared across TTT-style updates on CUT3R.

\begin{table*}[t]
\centering
\caption{\textbf{Pillar decoupling ablation.} We isolate each control of ReCal3R. \emph{ReCal3R w/ TTT3R rate} replaces our uncalibrated rate with TTT3R's alignment rate while keeping reliability calibration (state-side \emph{whether}); \emph{ReCal3R w/o reliability} keeps our uncalibrated rate but disables reliability calibration (observation-side \emph{how much}). Results are reported on camera pose estimation (ScanNet-50), video depth estimation (Bonn, scale-invariant), and 3D reconstruction (7-Scenes). For 7-Scenes, we report mean metrics only. Best in \textbf{bold}, second best \underline{underlined}.}
\label{tab:pillar-ablation}
\scriptsize
\setlength{\tabcolsep}{3.5pt}

\begin{subtable}{\textwidth}
\centering
\caption{Camera pose estimation on ScanNet-50.}
\begin{tabular*}{\textwidth}{@{\extracolsep{\fill}}l ccc ccc ccc}
\toprule
& \multicolumn{3}{c}{300 Frames} & \multicolumn{3}{c}{500 Frames} & \multicolumn{3}{c}{1000 Frames} \\
\cmidrule(lr){2-4}\cmidrule(lr){5-7}\cmidrule(lr){8-10}
Method & ATE $\downarrow$ & RPE$_{\mathrm{t}}$ $\downarrow$ & RPE$_{\mathrm{r}}$ $\downarrow$
       & ATE $\downarrow$ & RPE$_{\mathrm{t}}$ $\downarrow$ & RPE$_{\mathrm{r}}$ $\downarrow$
       & ATE $\downarrow$ & RPE$_{\mathrm{t}}$ $\downarrow$ & RPE$_{\mathrm{r}}$ $\downarrow$ \\
\midrule
CUT3R & 0.4863 & \textbf{0.0421} & \underline{1.1472} & 0.6394 & \underline{0.0343} & 1.0332 & 0.7865 & \textbf{0.0237} & 0.8429 \\
TTT3R & 0.1940 & \underline{0.0446} & 1.3822 & 0.2486 & \textbf{0.0329} & \textbf{0.8801} & 0.4230 & \underline{0.0244} & 0.8571 \\
ReCal3R w/ TTT3R rate & \textbf{0.1651} & 0.0534 & 1.1648 & \underline{0.1698} & 0.0346 & \underline{1.0154} & \underline{0.2431} & 0.0247 & \underline{0.8044} \\
ReCal3R w/o reliability & \underline{0.1729} & 0.0518 & 1.2145 & 0.1767 & 0.0356 & 1.3171 & 0.2502 & 0.0247 & \textbf{0.7592} \\
\textbf{ReCal3R} & 0.1739 & 0.0593 & \textbf{0.9008} & \textbf{0.1627} & 0.0377 & 1.0890 & \textbf{0.2106} & 0.0250 & 0.9069 \\
\bottomrule
\end{tabular*}
\end{subtable}

\vspace{0.8em}

\begin{subtable}{\textwidth}
\centering
\caption{Video depth estimation on Bonn (scale-invariant).}
\begin{tabular*}{0.78\textwidth}{@{\extracolsep{\fill}}l cc cc cc}
\toprule
& \multicolumn{2}{c}{300 Frames} & \multicolumn{2}{c}{500 Frames} & \multicolumn{2}{c}{700 Frames} \\
\cmidrule(lr){2-3}\cmidrule(lr){4-5}\cmidrule(lr){6-7}
Method & Rel $\downarrow$ & $\delta < 1.25$ $\uparrow$
       & Rel $\downarrow$ & $\delta < 1.25$ $\uparrow$
       & Rel $\downarrow$ & $\delta < 1.25$ $\uparrow$ \\
\midrule
CUT3R & 0.0681 & 0.9326 & 0.0664 & 0.9350 & 0.0697 & 0.9268 \\
TTT3R & 0.0581 & 0.9467 & 0.0594 & 0.9545 & 0.0632 & 0.9442 \\
ReCal3R w/ TTT3R rate & \underline{0.0544} & 0.9447 & \underline{0.0533} & 0.9562 & \underline{0.0562} & 0.9502 \\
ReCal3R w/o reliability & 0.0548 & \underline{0.9468} & 0.0535 & \textbf{0.9567} & 0.0565 & \underline{0.9508} \\
\textbf{ReCal3R} & \textbf{0.0535} & \textbf{0.9480} & \textbf{0.0512} & \underline{0.9562} & \textbf{0.0532} & \textbf{0.9515} \\
\bottomrule
\end{tabular*}
\end{subtable}

\vspace{0.8em}

\begin{subtable}{\textwidth}
\centering
\caption{3D reconstruction on 7-Scenes. We report mean metrics only.}
\begin{tabular*}{\textwidth}{@{\extracolsep{\fill}}l ccc ccc ccc}
\toprule
& \multicolumn{3}{c}{300 Frames} & \multicolumn{3}{c}{500 Frames} & \multicolumn{3}{c}{700 Frames} \\
\cmidrule(lr){2-4}\cmidrule(lr){5-7}\cmidrule(lr){8-10}
Method & Acc. $\downarrow$ & Comp. $\downarrow$ & NC $\uparrow$
       & Acc. $\downarrow$ & Comp. $\downarrow$ & NC $\uparrow$
       & Acc. $\downarrow$ & Comp. $\downarrow$ & NC $\uparrow$ \\
\midrule
CUT3R & 0.144 & 0.077 & 0.557 & 0.200 & 0.099 & 0.544 & 0.210 & 0.130 & 0.539 \\
TTT3R & 0.044 & 0.028 & 0.602 & 0.070 & 0.038 & 0.586 & 0.099 & 0.051 & 0.570 \\
ReCal3R w/ TTT3R rate & \underline{0.024} & \underline{0.025} & \underline{0.617} & \underline{0.035} & \underline{0.028} & \underline{0.613} & \underline{0.043} & \underline{0.030} & \underline{0.603} \\
ReCal3R w/o reliability & 0.028 & \textbf{0.024} & 0.613 & 0.040 & \underline{0.028} & 0.608 & 0.050 & 0.031 & 0.595 \\
\textbf{ReCal3R} & \textbf{0.023} & \textbf{0.024} & \textbf{0.619} & \textbf{0.029} & \textbf{0.026} & \textbf{0.614} & \textbf{0.034} & \textbf{0.028} & \textbf{0.608} \\
\bottomrule
\end{tabular*}
\end{subtable}
\end{table*}

\subsection{Exploration of Diagnostic Cues}
\label{sec:app-diagnostic-cues}

\paragraph{Single cue vs. fusion.}
Table~\ref{tab:cue-ablation} reports reliability estimates built from deviation alone, entropy alone, and their fusion. The fact that the two single-cue variants are already strong is itself informative: both deviation and entropy provide useful reliability evidence, so the reliability mechanism is not tied to a single accidental diagnostic. Deviation measures state-side displacement in the updated memory, while entropy reflects ambiguity in the attention distribution; when these cues co-vary, either cue can recover much of the benefit, which explains their close ATE, depth, and reconstruction results. We nevertheless retain the fusion because its role is not to win every individual entry, but to remove the need to choose a cue by hand and to remain stable under different failure modes. In particular, its conjunctive form defers to the more conservative cue when the two disagree, giving the most stable rotational accuracy on 300-frame ScanNet (RPE$_{\mathrm{rot}}=0.901$). The trade-off is that this conservatism can let a single cue match or slightly edge fusion on some metrics; we prefer the fused estimate for its more reliable default behavior.

\begin{table*}[t]
\centering
\caption{\textbf{Diagnostic-cue ablation.} Reliability estimates built from state deviation alone, attention entropy alone, and their fusion (full ReCal3R). Single-cue and fused estimates perform comparably across pose, depth, and reconstruction, while the fusion is hyperparameter-free and most stable in rotation. Results on camera pose (ScanNet-50), video depth (Bonn, scale-invariant), and 3D reconstruction (7-Scenes). Best in \textbf{bold}; ties after rounding are also bolded.}
\label{tab:cue-ablation}
\scriptsize
\setlength{\tabcolsep}{3.5pt}

\begin{subtable}{\textwidth}
\centering
\caption{Camera pose estimation on ScanNet-50.}
\begin{tabular*}{\textwidth}{@{\extracolsep{\fill}}l ccc ccc ccc}
\toprule
& \multicolumn{3}{c}{300 Frames} & \multicolumn{3}{c}{500 Frames} & \multicolumn{3}{c}{1000 Frames}\\
\cmidrule(lr){2-4}\cmidrule(lr){5-7}\cmidrule(lr){8-10}
Method & ATE $\downarrow$ & RPE$_{\mathrm{t}}$ $\downarrow$ & RPE$_{\mathrm{r}}$ $\downarrow$
       & ATE $\downarrow$ & RPE$_{\mathrm{t}}$ $\downarrow$ & RPE$_{\mathrm{r}}$ $\downarrow$
       & ATE $\downarrow$ & RPE$_{\mathrm{t}}$ $\downarrow$ & RPE$_{\mathrm{r}}$ $\downarrow$\\
\midrule
Deviation only & 0.1753 & \textbf{0.0574} & 1.2350 & 0.1627 & \textbf{0.0369} & \textbf{1.0544} & 0.2110 & 0.0251 & 0.7762\\
Entropy only & 0.1786 & 0.0604 & 1.3662 & \textbf{0.1605} & 0.0375 & 1.1796 & 0.2111 & 0.0253 & \textbf{0.7185}\\
\textbf{ReCal3R (fusion)} & \textbf{0.1739} & 0.0593 & \textbf{0.9008} & 0.1627 & 0.0377 & 1.0890 & \textbf{0.2106} & \textbf{0.0250} & 0.9069\\
\bottomrule
\end{tabular*}
\end{subtable}

\vspace{0.8em}

\begin{subtable}{\textwidth}
\centering
\caption{Video depth estimation on Bonn (scale-invariant).}
\begin{tabular*}{0.78\textwidth}{@{\extracolsep{\fill}}l cc cc cc}
\toprule
& \multicolumn{2}{c}{300 Frames} & \multicolumn{2}{c}{500 Frames} & \multicolumn{2}{c}{700 Frames}\\
\cmidrule(lr){2-3}\cmidrule(lr){4-5}\cmidrule(lr){6-7}
Method & Rel $\downarrow$ & $\delta < 1.25$ $\uparrow$
       & Rel $\downarrow$ & $\delta < 1.25$ $\uparrow$
       & Rel $\downarrow$ & $\delta < 1.25$ $\uparrow$\\
\midrule
Deviation only & \textbf{0.0534} & 0.9442 & 0.0513 & 0.9550 & 0.0537 & 0.9498\\
Entropy only & \textbf{0.0534} & 0.9442 & 0.0517 & 0.9555 & 0.0536 & 0.9508\\
\textbf{ReCal3R (fusion)} & 0.0535 & \textbf{0.9480} & \textbf{0.0512} & \textbf{0.9562} & \textbf{0.0532} & \textbf{0.9515}\\
\bottomrule
\end{tabular*}
\end{subtable}

\vspace{0.8em}

\begin{subtable}{\textwidth}
\centering
\caption{3D reconstruction on 7-Scenes. We report mean metrics only.}
\begin{tabular*}{\textwidth}{@{\extracolsep{\fill}}l ccc ccc ccc}
\toprule
& \multicolumn{3}{c}{300 Frames} & \multicolumn{3}{c}{500 Frames} & \multicolumn{3}{c}{700 Frames}\\
\cmidrule(lr){2-4}\cmidrule(lr){5-7}\cmidrule(lr){8-10}
Method & Acc. $\downarrow$ & Comp. $\downarrow$ & NC $\uparrow$
       & Acc. $\downarrow$ & Comp. $\downarrow$ & NC $\uparrow$
       & Acc. $\downarrow$ & Comp. $\downarrow$ & NC $\uparrow$\\
\midrule
Deviation only & \textbf{0.023} & 0.025 & \textbf{0.619} & \textbf{0.028} & 0.026 & 0.616 & \textbf{0.033} & 0.028 & \textbf{0.609}\\
Entropy only & \textbf{0.023} & \textbf{0.024} & 0.618 & 0.029 & \textbf{0.025} & \textbf{0.617} & 0.034 & \textbf{0.027} & 0.608\\
\textbf{ReCal3R (fusion)} & \textbf{0.023} & \textbf{0.024} & \textbf{0.619} & 0.029 & 0.026 & 0.614 & 0.034 & 0.028 & 0.608\\
\bottomrule
\end{tabular*}
\end{subtable}
\end{table*}

\paragraph{Fixed-anchor vs. moving-anchor state deviation.}
The reliability estimate uses a state-deviation cue to estimate whether a receiving state token has undergone excessive rewriting.
Using the row notation of Sec.~\ref{sec:preliminaries}, the fixed-anchor cue for state token $m$ is
\begin{equation}
    d_m^{\mathrm{init},(t)}
    =
    \left\|
    \mathbf{S}_m^{(t-1)} - \mathbf{S}_m^{(0)}
    \right\|_2 .
    \label{eq:app-delta-init}
\end{equation}
A natural alternative is to replace the fixed initial anchor with a moving reference, given by the running average of the token trajectory:
\begin{equation}
    \bar{\mathbf{S}}_m^{(t-1)}
    =
    \frac{1}{t}
    \sum_{\tau=0}^{t-1} \mathbf{S}_m^{(\tau)},
    \qquad
    d_m^{\mathrm{avg},(t)}
    =
    \left\|
    \mathbf{S}_m^{(t-1)} - \bar{\mathbf{S}}_m^{(t-1)}
    \right\|_2 .
    \label{eq:app-delta-avg}
\end{equation}
This variant keeps all other components unchanged and only replaces the displacement cue in the reliability estimation.

The comparison is motivated by a subtle design choice in streaming reconstruction. Although the model processes frames online, the reconstructed geometry and camera trajectory are maintained in a fixed reference frame rather than in a coordinate system that should freely drift with the state trajectory.

\begin{table*}[t]
    \centering
    \scriptsize
    \setlength{\tabcolsep}{3.5pt}
    \caption{\textbf{State-deviation cue ablation.} We compare the proposed first-state anchor with a running-average anchor across different sequence lengths. 7-Scenes reports Acc./Comp./NC, and ScanNet reports ATE.}
    \label{tab:ablation-state-drift}
    \resizebox{\textwidth}{!}{%
    \begin{tabular}{l!{\vrule}ccc|ccc|ccc!{\vrule}c|c|c}
        \toprule
        \multirow{3}{*}[-0.8ex]{Method}
        & \multicolumn{9}{c|}{7-Scenes}
        & \multicolumn{3}{c}{ScanNet} \\
        \cmidrule(lr){2-10}
        \cmidrule(l){11-13}
        & \multicolumn{3}{c|}{300 Frames}
        & \multicolumn{3}{c|}{500 Frames}
        & \multicolumn{3}{c|}{700 Frames}
        & 300 Frames
        & 500 Frames
        & 1000 Frames \\
        \cmidrule(lr){2-4}
        \cmidrule(lr){5-7}
        \cmidrule(lr){8-10}
        \cmidrule(lr){11-11}
        \cmidrule(lr){12-12}
        \cmidrule(l){13-13}
        & Acc. $\downarrow$
        & Comp. $\downarrow$
        & NC $\uparrow$
        & Acc. $\downarrow$
        & Comp. $\downarrow$
        & NC $\uparrow$
        & Acc. $\downarrow$
        & Comp. $\downarrow$
        & NC $\uparrow$
        & ATE $\downarrow$
        & ATE $\downarrow$
        & ATE $\downarrow$ \\
        \midrule
        ReCal3R w/ running avg.
        & 0.023
        & 0.024
        & 0.619
        & 0.029
        & 0.026
        & \textbf{0.617}
        & 0.036
        & 0.028
        & 0.606
        & 0.1829
        & 0.1674
        & 0.2151 \\
        \textbf{ReCal3R}
        & 0.023
        & 0.024
        & 0.619
        & 0.029
        & 0.026
        & 0.614
        & \textbf{0.034}
        & 0.028
        & \textbf{0.608}
        & \textbf{0.1739}
        & \textbf{0.1627}
        & \textbf{0.2106} \\
        \bottomrule
    \end{tabular}
    }
\end{table*}

Therefore, the diagnostic cue should preserve information about accumulated departure from the initial reference. The fixed-anchor deviation in Eq.~\ref{eq:app-delta-init} serves this purpose: it measures how far a token has been rewritten from its initial reference and therefore exposes long stream state deviation. In contrast, the running-average variant in Eq.~\ref{eq:app-delta-avg} uses a moving anchor that is itself affected by previous updates. As a result, gradual state deviation can be absorbed into the running average, making the token appear locally stable even when it has accumulated a large deviation from the original reconstruction reference.

We report this ablation on relative pose estimation and multi-view reconstruction in Tab.~\ref{tab:ablation-state-drift}. The running-average variant is consistently inferior to the fixed-anchor design. This suggests that, for streaming reconstruction over long streams, the diagnostic cue should not only capture local fluctuation around the token's recent trajectory, but should also retain a fixed reference for detecting accumulated state deviation. The results support our choice of using $\mathbf{S}_m^{(0)}$ as the state-side anchor in the reliability estimate. Because the global reconstruction remains expressed in the first-frame coordinate system, $\mathbf{S}_m^{(0)}$ provides the most faithful reference for measuring actual geometric deviation rather than deviation from a moving internal baseline. It is also more efficient: unlike the running average, it avoids per-frame trajectory aggregation and therefore does not incur the accompanying FPS drop.

\subsection{Anatomy of the Uncalibrated Rate}
\label{sec:app-beta-ablation}

\textbf{Anatomy of the uncalibrated rate.} Following the per-token convention of Sec.~\ref{sec:uncalibrated-rate}, we omit the token and time indices and write the uncalibrated learning rate as $\tilde\beta=g \cdot r \cdot \phi(h)=g \cdot r \cdot \exp(-h)$. Here, $g$ is the alignment gate inherited from TTT3R, $r$ is the state reconstruction residual, and $\phi(h)=\exp(-h)$ is the update pressure attenuation. To isolate the latter two factors, we form a residual-only variant $\tilde\beta=g \cdot r$ and a pressure-only variant $\tilde\beta=g \cdot \exp(-h)$, each retaining the alignment gate but removing one modulator. As described in Sec.~\ref{sec:uncalibrated-rate}, the residual suppresses writes from redundant frames that carry little unexplained content, while update pressure throttles tokens that have absorbed disproportionate recent evidence, preserving capacity for under-written slots.

Table~\ref{tab:beta-ablation} shows that removing either factor degrades performance across pose, depth, and reconstruction. On 1000-frame ScanNet, ATE rises from $0.250$ (full $\tilde\beta$) to $0.322$ and $0.327$ when only the residual or only update pressure is kept; the same ordering holds on 7-Scenes ($0.050\!\to\!0.079$ and $0.078$ at 700 frames) and on Bonn across all horizons. Notably, neither single factor dominates the other, indicating that they address non-interchangeable aspects of write quality rather than two estimates of the same quantity.

This pattern is the empirical signature of the multiplicative design in Eq.~\ref{eq:uncalibrated-rate}. Because the factors combine as a product, either one can independently veto an unsafe write: a redundant frame attenuates $r$ and suppresses the update globally, while a saturated token attenuates $\exp(-h)$ and suppresses its own update even on an informative frame. Dropping a factor removes one veto path, which is why no single-factor variant recovers the full rate.

\begin{table*}[t]
\centering
\caption{\textbf{Uncalibrated-rate ablation.} We ablate the state reconstruction residual and update pressure attenuation inside $\tilde\beta$ while retaining the inherited alignment gate $g$. \emph{Residual only} uses $g\cdot r$, \emph{pressure only} uses $g\cdot\exp(-h)$, and full $\tilde\beta$ uses $g\cdot r\cdot\exp(-h)$. Results are reported on camera pose estimation (ScanNet-50), video depth estimation (Bonn, scale-invariant), and 3D reconstruction (7-Scenes). For 7-Scenes, we report mean metrics only. Best in \textbf{bold}, second best \underline{underlined}.}
\label{tab:beta-ablation}
\scriptsize
\setlength{\tabcolsep}{3.5pt}

\begin{subtable}{\textwidth}
\centering
\caption{Camera pose estimation on ScanNet-50.}
\begin{tabular*}{\textwidth}{@{\extracolsep{\fill}}l ccc ccc ccc}
\toprule
& \multicolumn{3}{c}{300 Frames} & \multicolumn{3}{c}{500 Frames} & \multicolumn{3}{c}{1000 Frames} \\
\cmidrule(lr){2-4}\cmidrule(lr){5-7}\cmidrule(lr){8-10}
Method & ATE $\downarrow$ & RPE$_{\mathrm{t}}$ $\downarrow$ & RPE$_{\mathrm{r}}$ $\downarrow$
       & ATE $\downarrow$ & RPE$_{\mathrm{t}}$ $\downarrow$ & RPE$_{\mathrm{r}}$ $\downarrow$
       & ATE $\downarrow$ & RPE$_{\mathrm{t}}$ $\downarrow$ & RPE$_{\mathrm{r}}$ $\downarrow$ \\
\midrule
Residual only ($g\cdot r$) & 0.1804 & \underline{0.0477} & \underline{1.1208} & 0.2162 & \underline{0.0332} & \textbf{0.9099} & \underline{0.3221} & \underline{0.0248} & 0.8846 \\
Pressure only ($g\cdot\exp(-h)$) & \underline{0.1737} & \textbf{0.0456} & \textbf{1.0831} & \underline{0.2097} & \textbf{0.0330} & \underline{0.9615} & 0.3273 & 0.0248 & \underline{0.7935} \\
Full $\tilde\beta$ & \textbf{0.1729} & 0.0518 & 1.2145 & \textbf{0.1767} & 0.0356 & 1.3171 & \textbf{0.2502} & \textbf{0.0247} & \textbf{0.7592} \\
\bottomrule
\end{tabular*}
\end{subtable}

\vspace{0.8em}

\begin{subtable}{\textwidth}
\centering
\caption{Video depth estimation on Bonn (scale-invariant).}
\begin{tabular*}{0.78\textwidth}{@{\extracolsep{\fill}}l cc cc cc}
\toprule
& \multicolumn{2}{c}{300 Frames} & \multicolumn{2}{c}{500 Frames} & \multicolumn{2}{c}{700 Frames} \\
\cmidrule(lr){2-3}\cmidrule(lr){4-5}\cmidrule(lr){6-7}
Method & Rel $\downarrow$ & $\delta < 1.25$ $\uparrow$
       & Rel $\downarrow$ & $\delta < 1.25$ $\uparrow$
       & Rel $\downarrow$ & $\delta < 1.25$ $\uparrow$ \\
\midrule
Residual only ($g\cdot r$) & \underline{0.0557} & \textbf{0.9481} & \underline{0.0557} & \textbf{0.9572} & \underline{0.0590} & \underline{0.9502} \\
Pressure only ($g\cdot\exp(-h)$) & 0.0561 & \underline{0.9469} & 0.0560 & 0.9560 & 0.0602 & 0.9489 \\
Full $\tilde\beta$ & \textbf{0.0548} & 0.9468 & \textbf{0.0535} & \underline{0.9567} & \textbf{0.0565} & \textbf{0.9508} \\
\bottomrule
\end{tabular*}
\end{subtable}

\vspace{0.8em}

\begin{subtable}{\textwidth}
\centering
\caption{3D reconstruction on 7-Scenes. We report mean metrics only.}
\begin{tabular*}{\textwidth}{@{\extracolsep{\fill}}l ccc ccc ccc}
\toprule
& \multicolumn{3}{c}{300 Frames} & \multicolumn{3}{c}{500 Frames} & \multicolumn{3}{c}{700 Frames} \\
\cmidrule(lr){2-4}\cmidrule(lr){5-7}\cmidrule(lr){8-10}
Method & Acc. $\downarrow$ & Comp. $\downarrow$ & NC $\uparrow$
       & Acc. $\downarrow$ & Comp. $\downarrow$ & NC $\uparrow$
       & Acc. $\downarrow$ & Comp. $\downarrow$ & NC $\uparrow$ \\
\midrule
Residual only ($g\cdot r$) & 0.034 & \underline{0.025} & 0.608 & 0.059 & 0.036 & 0.595 & 0.079 & 0.042 & 0.583 \\
Pressure only ($g\cdot\exp(-h)$) & \underline{0.033} & 0.026 & \underline{0.611} & \underline{0.057} & \underline{0.031} & \underline{0.597} & \underline{0.078} & \underline{0.038} & \underline{0.584} \\
Full $\tilde\beta$ & \textbf{0.028} & \textbf{0.024} & \textbf{0.613} & \textbf{0.040} & \textbf{0.028} & \textbf{0.608} & \textbf{0.050} & \textbf{0.031} & \textbf{0.595} \\
\bottomrule
\end{tabular*}
\end{subtable}
\end{table*}

\subsection{Sensitivity to the Conservative Fallback Rate}
\label{sec:app-beta-base}

The conservative fallback rate $\beta^{\mathrm{base}}$ controls the residual plasticity of low-reliability state tokens. In Eq.~\ref{eq:calibrated-rate}, a smaller $\beta^{\mathrm{base}}$ makes the update increasingly rely on the reliability calibration, and degenerates to freezing low-reliability tokens when $\beta^{\mathrm{base}}=0$. In contrast, a larger $\beta^{\mathrm{base}}$ allows low-reliability tokens to keep absorbing information, but also weakens the protective effect of the reliability calibration by assigning non-negligible updates even when the state-side diagnosis is uncertain. We therefore evaluate different choices of $\beta^{\mathrm{base}}$ while keeping all other components fixed.

\begin{table*}[htbp]
    \centering
    \small
    \setlength{\tabcolsep}{4pt}
    \caption{\textbf{Sensitivity to the conservative fallback rate.}
    Camera pose estimation results on ScanNet-50 sequences under different values of $\beta^{\mathrm{base}}$.
    ATE, RPE$_{\mathrm{trans}}$, and RPE$_{\mathrm{rot}}$ are reported for 300-, 500-, and 1000-frame input streams.
    Lower is better for all metrics.}
    \label{tab:beta-base-sensitivity}
    \begin{tabular}{c|ccc|ccc|ccc}
        \toprule
        \multirow{2}{*}{$\beta^{\mathrm{base}}$}
        & \multicolumn{3}{c|}{300 Frames}
        & \multicolumn{3}{c|}{500 Frames}
        & \multicolumn{3}{c}{1000 Frames} \\
        \cmidrule(lr){2-4}
        \cmidrule(lr){5-7}
        \cmidrule(lr){8-10}
        & ATE $\downarrow$
        & RPE$_{\mathrm{trans}}$ $\downarrow$
        & RPE$_{\mathrm{rot}}$ $\downarrow$
        & ATE $\downarrow$
        & RPE$_{\mathrm{trans}}$ $\downarrow$
        & RPE$_{\mathrm{rot}}$ $\downarrow$
        & ATE $\downarrow$
        & RPE$_{\mathrm{trans}}$ $\downarrow$
        & RPE$_{\mathrm{rot}}$ $\downarrow$ \\
        \midrule
        $0$
        & 0.362 & 0.125 & 6.711
        & 0.281 & 0.080 & 3.268
        & 0.244 & 0.044 & 1.721 \\
        $0.01$
        & 0.306 & 0.107 & 4.559
        & 0.223 & 0.063 & 1.893
        & \underline{0.200} & 0.037 & 1.273 \\
        $0.05$
        & 0.212 & 0.073 & 2.071
        & \underline{0.167} & 0.044 & 1.279
        & \textbf{0.190} & 0.027 & \textbf{0.818} \\
        $0.1$
        & \underline{0.174} & \underline{0.059} & \textbf{0.901}
        & \textbf{0.163} & \underline{0.038} & \underline{1.089}
        & 0.211 & \textbf{0.025} & 0.867 \\
        $0.2$
        & \textbf{0.164} & \textbf{0.048} & \underline{1.098}
        & 0.193 & \textbf{0.033} & \textbf{0.867}
        & 0.308 & \underline{0.025} & \underline{0.853} \\
        \bottomrule
    \end{tabular}
\end{table*}

As shown in Tab.~\ref{tab:beta-base-sensitivity}, setting $\beta^{\mathrm{base}}=0$ is consistently suboptimal. This indicates that a low reliability score should not completely disable state writing: even when the current update is considered unreliable, a small fallback path is still needed to maintain minimal plasticity and accumulate evidence for subsequent frames. Meanwhile, the preferred fallback strength depends on the sequence horizon. On shorter sequences, a stronger fallback can be beneficial because the fixed-size state has not yet accumulated severe update pressure over long streams, and allowing low-reliability tokens to keep updating improves short-term adaptation. However, as the input stream becomes longer, overly large fallback rates become less favorable, since they bypass the reliability calibration and allow uncertain tokens to receive repeated non-negligible writes. This accumulated effect is particularly visible on long sequences, where excessive fallback rates degrade global trajectory accuracy.

Overall, the results reveal a clear stability–plasticity trade-off.
A zero fallback is too conservative, while an overly large fallback weakens the purpose of reliability-calibrated gating under accumulation over long streams. The default value $\beta^{\mathrm{base}}=0.1$ provides a strong overall trade-off across sequence lengths and pose metrics, and is used in all main experiments.

\section{More Results}
\label{sec:app:more-results}

\subsection{Efficiency and effectiveness.}

\begin{figure*} [htbp]
    \centering
    \includegraphics[width=1\linewidth]{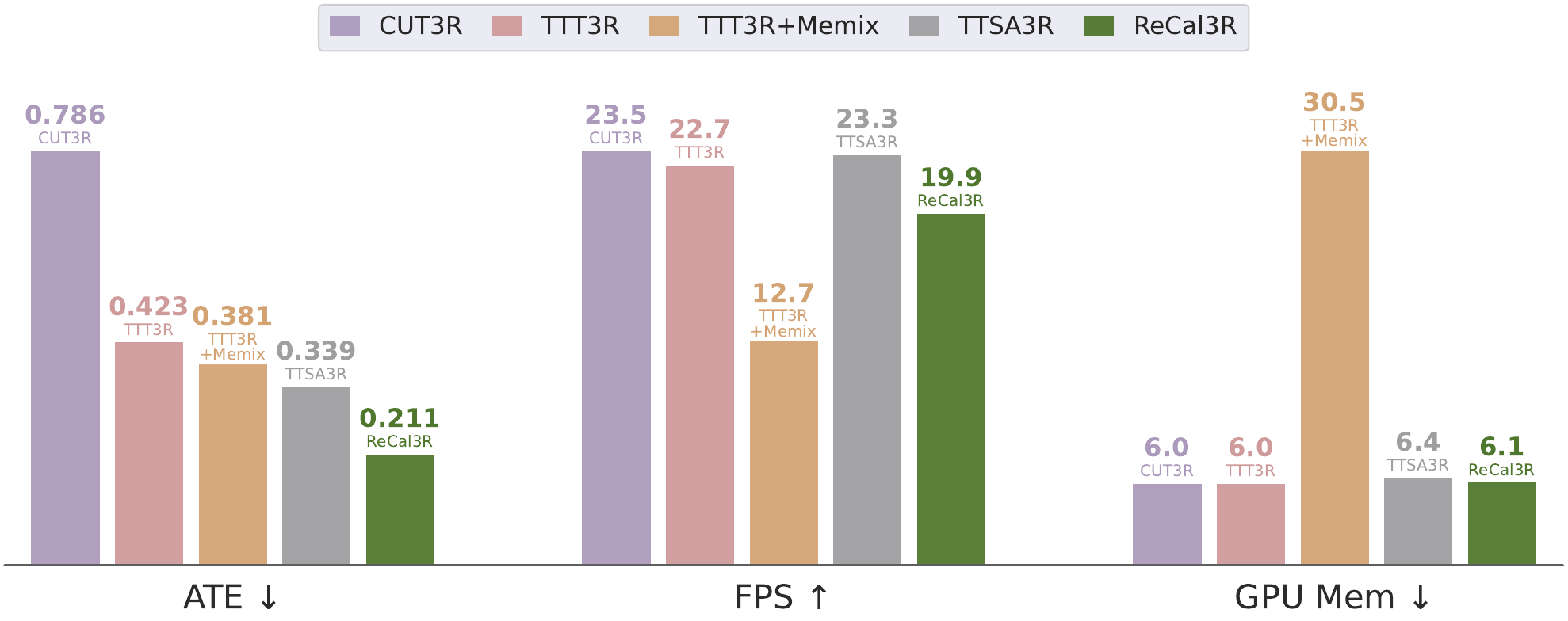}
    \caption{Streaming inference comparison. On 1,000-frame ScanNet sequences, ReCal3R achieves the lowest ATE among baselines while maintaining competitive FPS and CUT3R-level GPU memory usage.}
    \label{fig:teaser}
\end{figure*}

As shown in Fig.~\ref{fig:teaser}, ReCal3R achieves a \(3.7\times\) ATE reduction over CUT3R on 1{,}000-frame ScanNet sequences, while maintaining comparable FPS and nearly identical GPU memory usage. This confirms that ReCal3R improves long stream reconstruction accuracy without increasing the streaming state size or computational budget, achieving both effective and efficient online reconstruction.

\subsection{Camera Pose Estimation}
\label{sec:app:more-results-camera-pose}

\begin{figure*}[t]
    \centering
    \includegraphics[width=\linewidth]{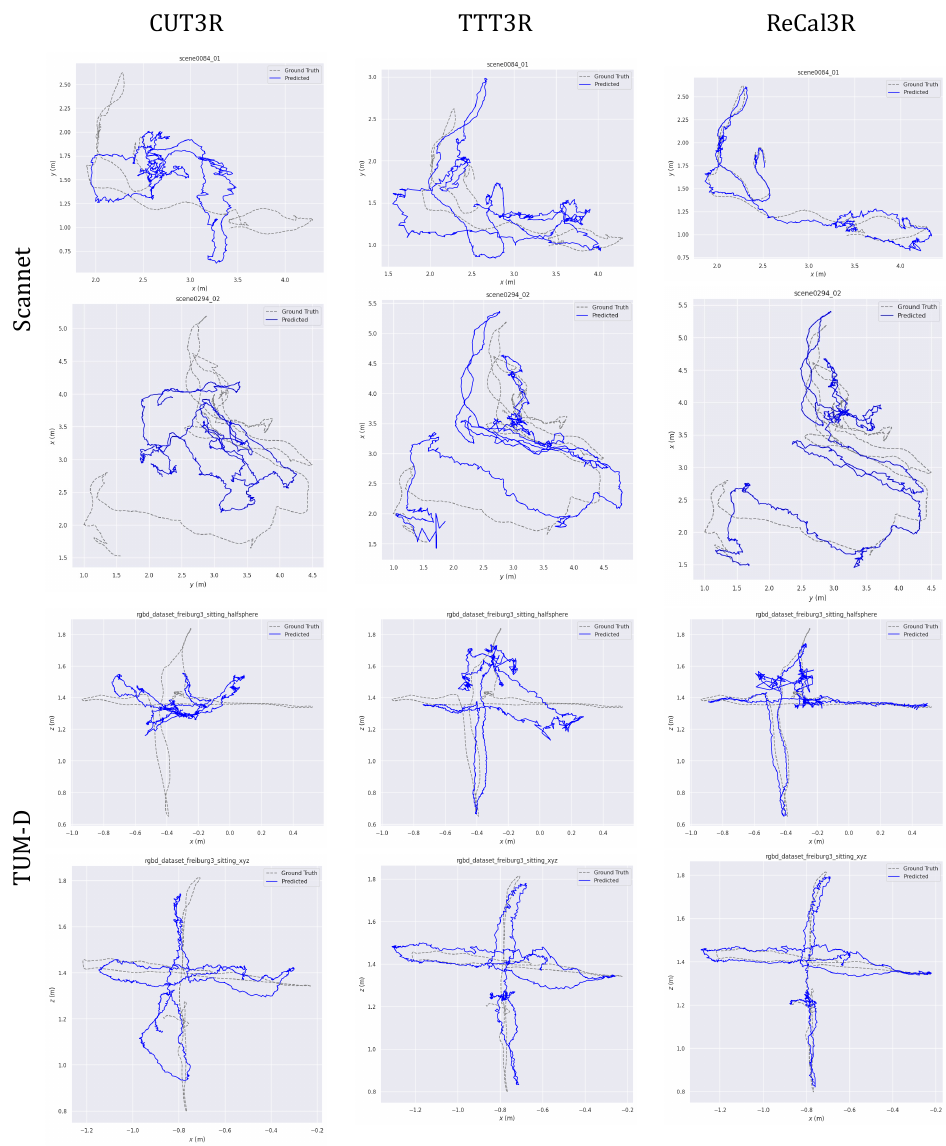}
    \caption{\textbf{Qualitative comparison of camera pose estimation.} ReCal3R produces camera trajectories that more faithfully follow the ground truth over long sequences, with reduced drift accumulation compared with CUT3R.}
    \label{fig:app-relpose-viz}
\end{figure*}

We provide additional qualitative comparisons of camera trajectory estimation in Fig.~\ref{fig:app-relpose-viz}. Consistent with the quantitative results in the main paper, ReCal3R produces trajectories that remain substantially closer to the ground truth over long streams, while CUT3R exhibits more noticeable drift accumulation. This advantage follows directly from our reliability-calibrated state-writing formulation. By decoupling \emph{whether} a state token should be updated from \emph{how much} new evidence it should absorb, ReCal3R prevents degraded or ambiguous state tokens from repeatedly incorporating unreliable observations. At the same time, reliable tokens can still adapt to genuinely novel content through the uncalibrated (candidate) learning rate. As a result, ReCal3R better preserves the geometric consistency of the recurrent state throughout the sequence, yielding more accurate and robust camera pose estimation with only lightweight, plug-and-play modifications to the CUT3R transition rule.

\subsection{Additional Qualitative Results on DL3DV}
\label{sec:app:more-results-dl3dv}

\begin{figure*}[t]
    \centering
    \includegraphics[width=\linewidth]{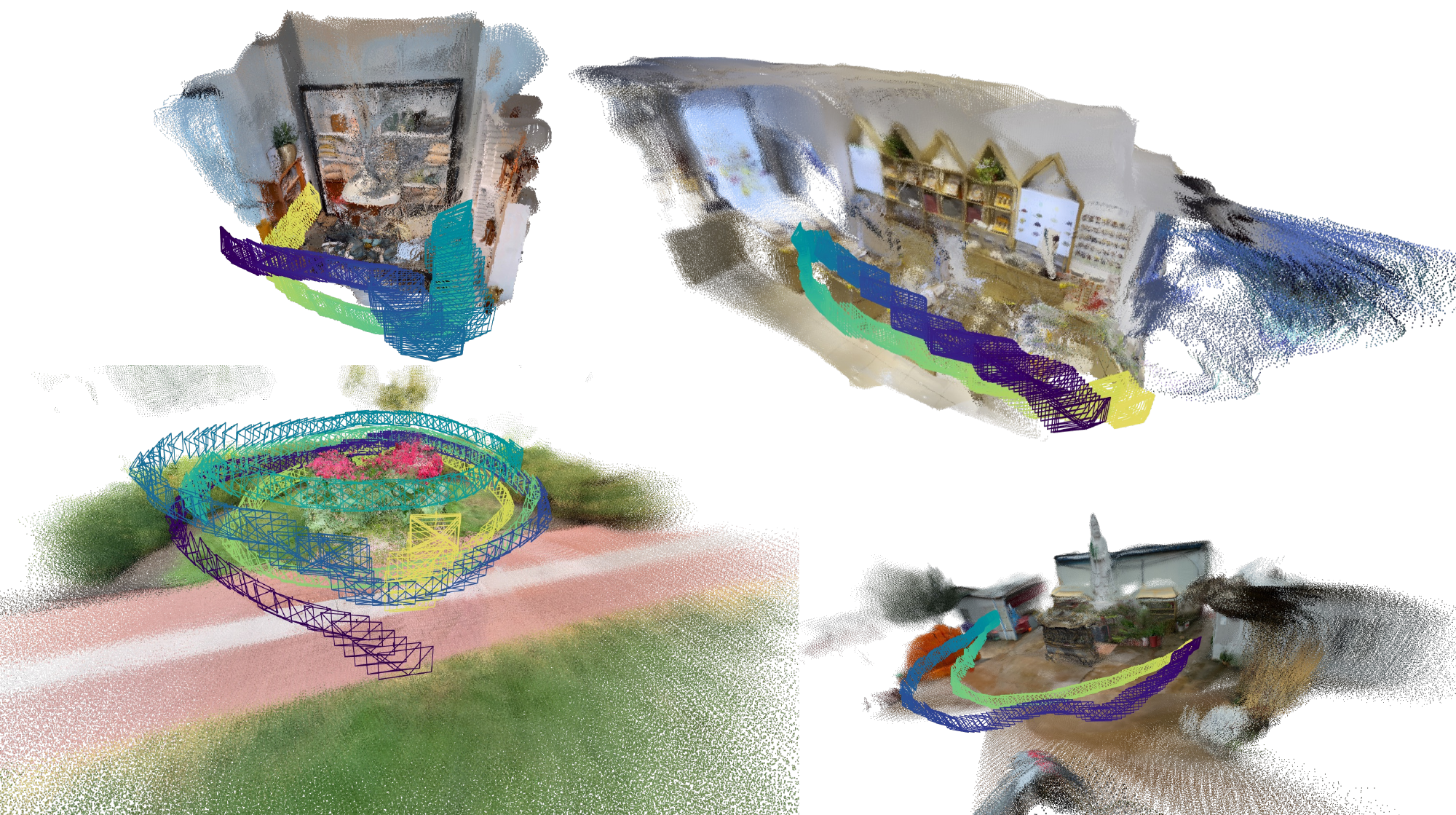}
    \caption{\textbf{Additional qualitative results on DL3DV.} The visualizations are produced directly from uninterrupted streaming inference over the full sequence, without applying any state reset or sequence-wise reinitialization.}
    \label{fig:app-dl3dv-viz}
\end{figure*}

We provide additional qualitative visualizations on representative DL3DV sequences in Fig.~\ref{fig:app-dl3dv-viz}. Importantly, these results are obtained in a fully streaming manner without any state reset or sequence-wise reinitialization. All visualizations are shown directly from the model outputs over the entire sequence, thereby reflecting the unmodified long stream behavior of the method.

\end{document}